\documentclass[10pt,twocolumn,letterpaper]{article}

\usepackage{wacv}
\usepackage{times}
\usepackage{epsfig}
\usepackage{graphicx}
\usepackage{amsmath}
\usepackage{amssymb}

\usepackage[table]{xcolor}
\usepackage{mathtools}
\usepackage{amsmath}
\usepackage{algorithm}
\usepackage[noend]{algpseudocode}
\usepackage{amssymb}
\usepackage{varwidth}
\usepackage{multirow}
\usepackage{pifont}
\usepackage{subfigure}
\usepackage{kotex}
\usepackage{array}
\usepackage{graphicx}
\newcommand{\cmark}{\ding{52}}
\newcommand{\xmark}{\ding{56}}
\usepackage{booktabs}

\def\wacvPaperID{0} 

\wacvfinalcopy 

\ifwacvfinal
\def\assignedStartPage{1} 
\fi

\ifwacvfinal
\usepackage[breaklinks=true,bookmarks=false]{hyperref}
\else
\usepackage[pagebackref=true,breaklinks=true,colorlinks,bookmarks=false]{hyperref}
\fi

\ifwacvfinal
\else
\fi

\begin{document}

\title{Continual Unsupervised Domain Adaptation for Semantic Segmentation}

\author{Joonhyuk Kim\textsuperscript{1}\thanks{Equal contribution.} \hspace{0.5cm} Sahng-Min Yoo\textsuperscript{2}\footnotemark[1] \hspace{0.5cm} Gyeong-Moon Park\textsuperscript{3} \hspace{0.5cm} Jong-Hwan Kim\textsuperscript{2}\\
${}^{1}$Seoul National University, Seoul, Republic of Korea \\ 
${}^{2}$KAIST, Daejeon, Republic of Korea \\
${}^{3}$Kyung Hee University, Yongin, Republic of Korea\\
{\tt\small ${}^{1}$kjh42551@snu.ac.kr, ${}^{2}$\{smyoo, johkim\}@rit.kaist.ac.kr, ${}^{3}$gmpark@khu.ac.kr}
}

\maketitle

\begin{abstract}
   Unsupervised Domain Adaptation (UDA) for semantic segmentation has been favorably applied to real-world scenarios in which pixel-level labels are hard to be obtained. In most of the existing UDA methods, all target data are assumed to be introduced simultaneously. Yet, the data are usually presented sequentially in the real world. 
   Moreover, Continual UDA, which deals with more practical scenarios with multiple target domains in the continual learning setting, has not been actively explored.
   In this light, we propose Continual UDA for semantic segmentation based on a newly designed Expanding Target-specific Memory (ETM) framework. Our novel ETM framework contains Target-specific Memory (TM) for each target domain to alleviate catastrophic forgetting. Furthermore, a proposed Double Hinge Adversarial (DHA) loss leads the network to produce better UDA performance overall. Our design of the TM and training objectives let the semantic segmentation network adapt to the current target domain while preserving the knowledge learned on previous target domains. The model with the proposed framework outperforms other state-of-the-art models in continual learning settings on standard benchmarks such as GTA5, SYNTHIA, CityScapes, IDD, and Cross-City datasets. The source code is available at \url{https://github.com/joonh-kim/ETM}.
\end{abstract}

\section{Introduction}
\label{sec:intro}

Deep learning-based approaches have shown remarkable improvements in semantic segmentation tasks via supervised learning \cite{long2015fully, ronneberger2015u, noh2015learning, chen2017deeplab, he2017mask, chen2017rethinking, lin2017feature, jegou2017one, yu2018bisenet, yu2018learning}. However, pixel-level labeling for datasets containing enormous real-world images usually requires a high cost of time and labor \cite{cordts2016cityscapes, varma2019idd, chen2017no, yu2020bdd100k, MVD2017}. 
Typically, such pixel-level labels can be automatically generated from synthetically rendered images \cite{richter2016playing, ros2016synthia}, however, a domain discrepancy between synthetic and real-world images is problematic. Therefore, many Unsupervised Domain Adaptation (UDA) techniques \cite{hoffman2016fcns, tsai2018learning, vu2019advent, chen2019domain, zhang2019category, wang2020differential}, which aim to adapt the network trained on synthetic images to real images, have been introduced to solve the domain discrepancy problem. 

Most of the existing UDA methods, however, consider an impractical scenario which only focuses on a single-target setting \cite{ganin2016domain, tzeng2017adversarial, deng2019cluster, hoffman2016fcns, tsai2018learning, vu2019advent, chen2019domain, zhang2019category, wang2020differential}. In the real world, there can be multiple target domains \cite{liu2020open, park2020discover}, and such domains may not be even introduced at once \cite{hoffman2014continuous, bobu2018adapting, wulfmeier2018incremental}. To this end, in this paper, we consider a more realistic Continual UDA scenario. Under this setting, the network trained on a source domain aims to adapt to multiple target domains which are presented sequentially.

\begin{figure*}[!t]
\centering  
\includegraphics[width=\linewidth]{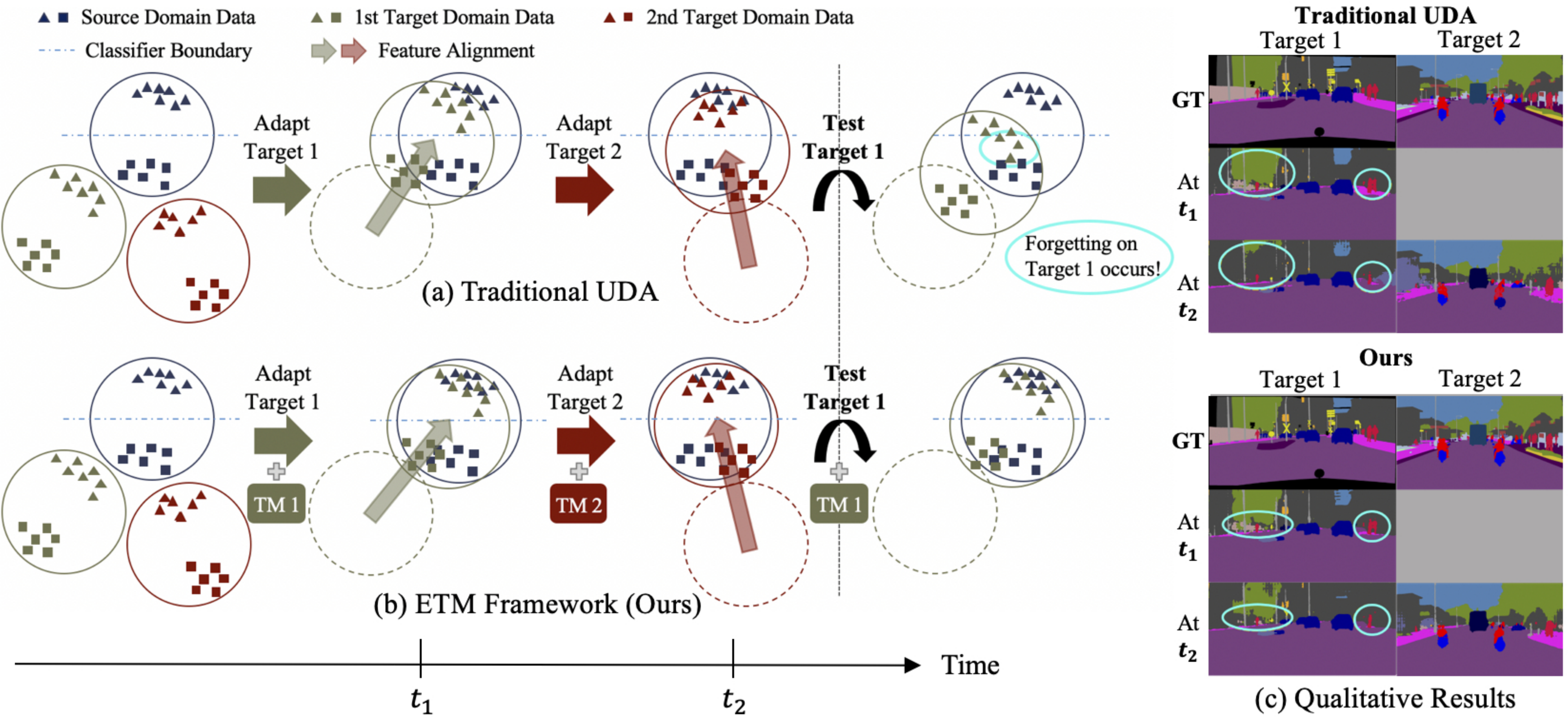}
\caption{(a) When the traditional UDA method is applied to the Continual UDA task, it suffers from catastrophic forgetting for the previously learned target domain (Target 1). (b) Our proposed ETM framework alleviates such forgetting by expanding a lightweight sub-memory called TM. (c) Qualitative results show that our framework actually mitigates forgetting on the previous target domain. AdaptSegNet \cite{tsai2018learning} is used as the baseline UDA method.}
\label{fig:intro}
\end{figure*}

Blind application of the existing UDA methods to this setting leads to sub-optimal results. We observe that notorious catastrophic forgetting \cite{mccloskey1989catastrophic, french1999catastrophic, mermillod2013stability, kirkpatrick2017overcoming} occurs for the previous target domain as the network is trained on the current target domain (see Fig. \ref{fig:intro}(c)).
Recently, Wu \etal \cite{wu2019ace} introduced a method for adapting to changing environments such as varying weather and lighting conditions, which can be viewed as Continual UDA for semantic segmentation. By using the style-memory of each environment, the method transfers the style of the source environment into that of each target environment. It achieves superior performance over the previous UDA methods under the specific setting. However, the experiments were conducted within one synthetic dataset \cite{ros2016synthia}, and accordingly, we observe inferior performance when applied to multiple real-world datasets (see Sec. \ref{sec:exp_sota}). We see that the style transfer is not enough to overcome the catastrophic forgetting problem when considerable domain discrepancy exists.

To this end, we propose a novel \textit{Expanding Target-specific Memory} (ETM) framework for Continual UDA for semantic segmentation on real-world datasets. In the framework, we introduce \textit{Target-specific Memory} (TM) for each target domain. Inspired by the previous works \cite{wang2017growing, yoon2017lifelong, park2020convolutional} in the continual learning field, it is considered that the constant capacity of the existing networks may not be enough to handle multiple target domains as it faces a huge domain discrepancy. The proposed framework is illustrated in Fig. \ref{fig:intro}(b). Specifically, a lightweight sub-memory called TM is initiated, trained, and stored for each target domain. Each TM contains unique information corresponding to each domain discrepancy by designing the TM in consideration of the structure, the forwarding path and the expanded location (see Section \ref{sec:TM}) .
When testing the network over the previous domains, the stored TM of the corresponding previous domain is used. In this way, the network overcomes the catastrophic forgetting problem. 
In addition, we design a \textit{Double Hinge Adversarial} (DHA) loss that enhances the overall UDA performance. By optimizing the DHA loss function, the segmentation network aligns the source and target domain data while considering geometric relations between them. We observe that the DHA loss is more suitable for the UDA objective. Without loss of generality, our framework can be applied to other adversarial learning-based UDA methods.

The main contribution of this paper is three-fold:
\begin{itemize}
	\item To the best of our knowledge, we address Continual UDA for semantic segmentation on real-world datasets for the first time, which considers more practical scenarios.
	\item We propose the ETM framework for Continual UDA. We deal with the catastrophic forgetting problem by expanding a little amount of model capacity (TM), which is the way that is firstly introduced in this field. Moreover, we propose the DHA loss function to enhance the performance of UDA with adversarial learning.
	\item We validate our framework by conducting experiments using two synthetic datasets (GTA5 \cite{richter2016playing}, SYNTHIA \cite{ros2016synthia}), and three real-world datasets (CityScapes \cite{cordts2016cityscapes}, IDD \cite{varma2019idd}, Cross-City \cite{chen2017no}) with large domain discrepancy. The model trained with the ETM framework outperforms other state-of-the-art models under the same conditions.
\end{itemize}

\section{Approach}
\label{sec:approach}

We first formalize the Continual UDA problem by defining the following notations. Let $\mathcal{S}=\{(x_1^s, y_1^s), \ldots,(x_{N_s}^s, y_{N_s}^s)\}$ denote the source domain data, which consists of $N_s$ images ($x_1^s, ... ,x_{N_s}^s$) and corresponding labels ($y_1^s, ... ,y_{N_s}^s$). Multiple $T$ target domains without any annotations are defined as $\{\mathcal{T}_i\}_{i=1}^{T}$. The $i$-th target domain data are defined as $\mathcal{T}_i=\{x_1^i, \ldots,x_{N_i}^i\}$, where the domain has $N_i$ images and $i \in \{1,\ldots,T\}$. 
Let $x$ and $y^s$ stand for an arbitrary image from any domains and its label if the image is drawn from the source domain, respectively. Then, $x \in \mathbb{R}^{3\times H\times W}$ and $y^s \in \mathbb{R}^{C\times H\times W}$, where $C$ is the number of classes, and $H$ and $W$ are height and width of the image. 
Moreover, we define arbitrary images drawn from the source and target domains as $x^s$ and $x^t$, respectively.

\begin{figure*}[!t]
\centering  
\subfigure[Proposed framework]{\includegraphics[width=0.81\linewidth]{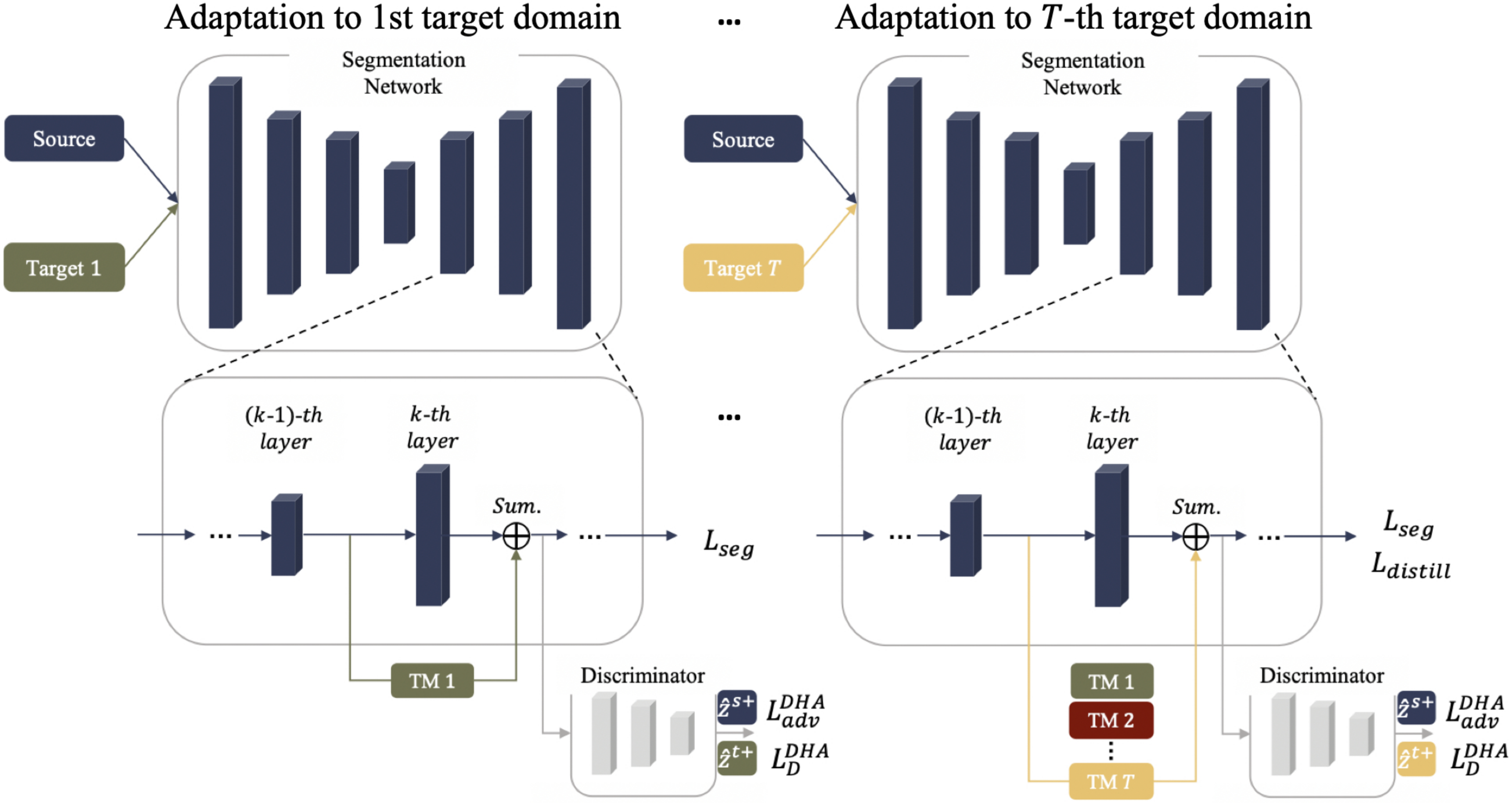}}
\subfigure[TM structure]{\includegraphics[width=0.185\linewidth]{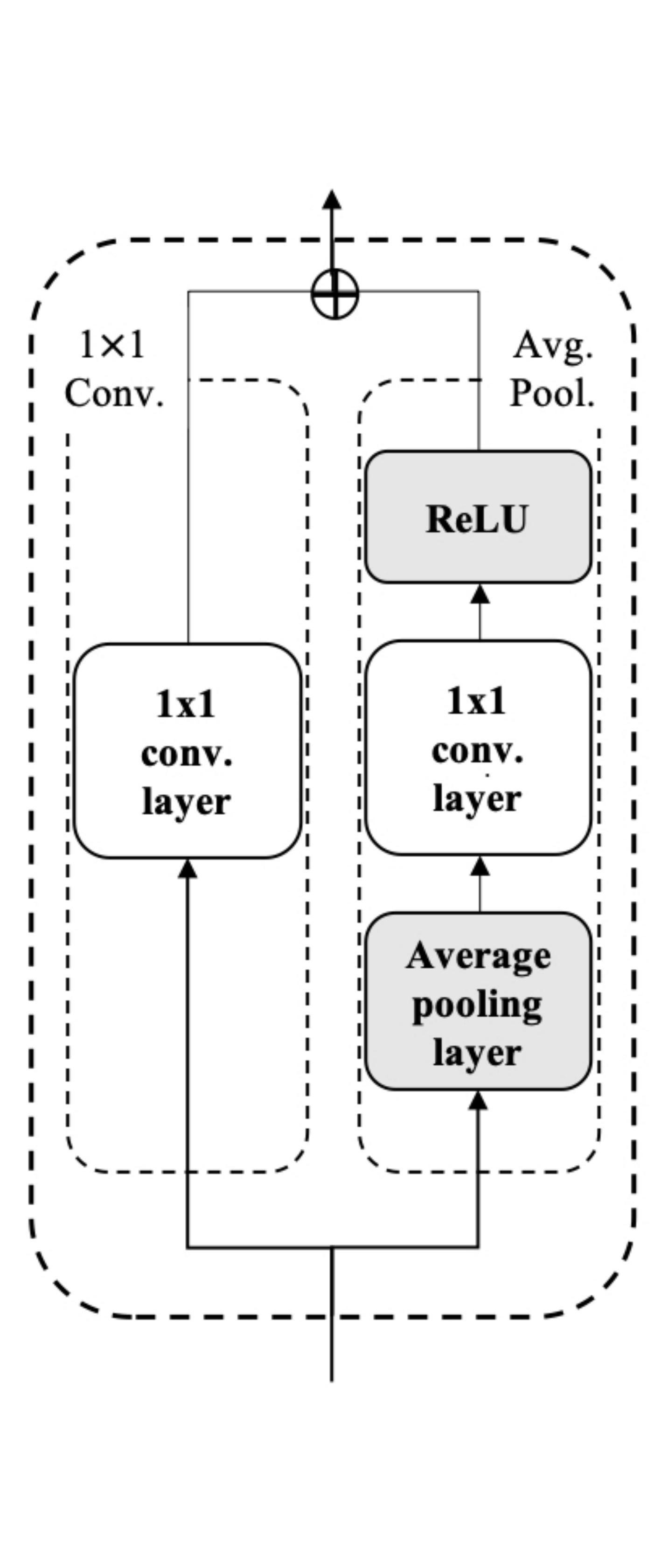}}
\caption{(a) The overall architecture of the ETM framework. The TM for each target domain is generated as the data appear sequentially. The discriminator tries to distinguish the source and target domain data, while the segmentation network and the TM try to fool the discriminator. When the $i$-th target domain data are introduced, corresponding TM is added right before the layer from the segmentation network entering into the discriminator. Note that $k$ is decided by the adversarial learning procedure of the backbone UDA method. (b) The detailed structure of the TM. Each TM consists of the $1\times1$ Conv. module that aims to extract the localized information, and the Avg. Pool module to extract the contextual information.}
\label{fig:main}
\end{figure*}

\subsection{Target-specific Memory}
\label{sec:TM}

We start with an idea that the network needs additional capacity in real-world scenarios where the number of target domains keeps increasing.
Thus, the added sub-memory needs to be stored and retrieved to reproduce its learned knowledge. 
There have already been precedents for expanding the network capacity in continual learning for the image classification task \cite{wang2017growing, yoon2017lifelong, park2020convolutional}. However, UDA for semantic segmentation differs from the image classification task in many aspects such as network structure, training objectives, and training strategies. Thus, adapting this idea to Continual UDA is not trivial. 

We first consider the structure of the TM. Above all, each TM should contain domain discrepancy information between the source domain and each target domain. Unlike in the image classification task, the added sub-memory cannot be a simple MLP network. In semantic segmentation, the ``accurate localization'' information can be contained by using the small receptive field with the small dilation value, and ``context assimilation'' information can be contained by considering the large receptive field with the large dilation value \cite{chen2017deeplab}. From Fig. \ref{fig:intro}(c), we observe that forgetting occurs for various objects from sidewalk to person. This indicates that there exists the domain discrepancy not only for the localized information but also for the contextual information. To this end, we design two modules for each TM. One for the localized information, and the other for the contextual information.

Moreover, the TM should be lightweight memory for the efficiency of the overall continual learning scheme. Thus, we design the sub-memory structure as follows: As shown in Fig. \ref{fig:main}(b), a $1\times1$ Conv. module ($1\times1$ convolution layer), which represents the module with the smallest receptive field. An Avg. Pool. module represents the module having the largest receptive field, which consists of an average pooling layer followed by a $1\times1$ convolution layer and the ReLU nonlinearity. Since the trainable parameters only exist in the $1\times1$ convolution layers, the TM is a fairly small-capacity network compared to the segmentation network.

The next step is for allowing each TM to contain each domain discrepancy information of the corresponding target domain during training. To do so, the knowledge, which is originally contained only in the segmentation network, should be divided up with the TM. Therefore, we let the TM be trained along with the segmentation network by attaching it to the network. In addition, to take advantage of high-level features, we expanded the TM in the top layer.

To be specific, when the input enters the TM it passes through the $1\times1$ Conv. and Avg. Pool. modules, respectively, and then the outputs are summed.
The output of the TM for arbitrary input $x$ is expressed as $TM(x)$ (Here, $TM$ indicates an arbitrary $i$-th TM, $TM_i$. We omit the subscription $i$ for simplicity.).
$TM(x)$ and the hidden state value of the segmentation network are added, and then pass through the next layers.
For example, the $(k-1)$-th hidden state value passes to the $k$-th layer and the TM, respectively, and then the added output passes through the $(k+1)$-th layer (see Fig. \ref{fig:main}(a)).
Let us express the output value as $f^{[:m]}(x)$ when the input passes through until the $m$-th layer of the segmentation network. Similarly, the output value when the input passes from the $(n+1)$-th layer of the segmentation network is expressed as $f^{[n:]}(x)$.
Then, the added output right before the $(k+1)$-th layer is represented as follows:
\begin{equation}
\begin{aligned}
& f^{[:k]+}(x) = f^{[:k]}(x) + TM(f^{[:(k-1)]}(x)).
\end{aligned}
\label{eq:f[:k]+}
\end{equation}
In addition, the final output passed through both the segmentation network and the TM is defined as follows:
\begin{equation}
\begin{multlined}
f^{+}(x) = f^{[k:]}(f^{[:k]+}(x)).
\end{multlined}
\label{eq:f+}
\end{equation}
In our framework, the source and target domain data are aligned via adversarial learning. The layer where the TM is added is decided by the adversarial learning procedure of the backbone UDA method. It can be multiple positions if the backbone method uses various features for a discriminator input. If the TM is added at the $k$-th layer, its hidden state value is used as the input of the discriminator.
In other words, $f^{[:k]+}(x)$ in Eq. (\ref{eq:f[:k]+}) becomes the input of the discriminator.

\subsection{Double Hinge Adversarial Loss}
\label{sec:DHA}

Typically, the UDA methods perform the alignment of the source and target domain data through adversarial learning with the discriminator \cite{tzeng2017adversarial, tsai2018learning, vu2019advent, wang2020differential} based on GAN \cite{goodfellow2014generative}. Since the segmentation network in UDA can correspond to the generator of GAN, the adversarial learning scheme has been simply adapted to the UDA field.
By minimizing the adversarial loss, the segmentation network is trained to align the source and target domain data, and at the same time, the discriminator is trained to distinguish them as the discriminator loss is minimized. There is a variation of GAN \cite{lim2017geometric} which considers the geometric relations between feature vectors in the hyperplane by using the Support Vector Machine (SVM) \cite{scholkopf2002learning}. In \cite{lim2017geometric}, the discriminator maximizes the minimum margin by value $1$ of the separable data in the feature space. This leads to stable training and solves the mode collapse problem in the image generation task.

However, the generator's task in GAN and that of the segmentation network in UDA are different, while the discriminator's role is in line with each other. Thus, we design a new adversarial loss that is customized for UDA.
In the feature space, the generator tries to move the fake feature vectors toward the real feature space so that they can be classified as the real feature vectors. This is because the real data are the absolute truth in GAN.
On the other hand, the target feature vectors in UDA do not have to imitate the source feature vectors blindly. Thus, the segmentation network can achieve its goal in two ways: by pulling the target features toward the source features or by pushing the source features toward the target features. Our suggestion is to use both ways.
Here, we do not update the segmentation network when the source features are already closer to the target side than the target features and vice versa, by using the ReLU nonlinearity (see Eq. (\ref{eq:DHA_adv})).

To this end, we propose the DHA loss for the segmentation network. This loss is minimized when the target features are extracted as if the input is from the source domain, and vice versa. The DHA loss replaces the loss functions for adversarial learning used in the backbone UDA method. Let $\hat{z}^s$ and $\hat{z}^t$ be the output values when $x^s$ and $x^t$ pass through the discriminator after the segmentation network, respectively. Then, $\hat{z}^s, \hat{z}^t \in \mathbb{R}^{h\times w}$, where $h$ and $w$ are height and width of each value, respectively. Then, our DHA loss is expressed as

\begin{equation}
\begin{aligned}
& L_{D}^{DHA}(\hat{z}^s, \hat{z}^t) = \sum_i^{h} \sum_j^{w} \big[(1 - \hat{z}^{s(i, j)})_{+} + (1 + \hat{z}^{t(i, j)})_{+} \big], \\
\end{aligned}
\label{eq:DHA_disc}
\end{equation}
\begin{equation}
\begin{aligned}
& L_{adv}^{DHA}(\hat{z}^s, \hat{z}^t) = \sum_i^{h} \sum_j^{w} (\hat{z}^{s(i, j)} - \hat{z}^{t(i, j)})_{+},
\end{aligned}
\label{eq:DHA_adv}
\end{equation}
where $(\cdot)_{+}=max(\, \cdot \,, 0)$.

In short, we propose an adversarial learning procedure customized for the UDA task.

\subsection{Expanding Target-specific Memory Framework}
\label{sec:ETM}

Even if the TM for the previous target domain remains frozen, however, the significant change in the segmentation network may aggravate the forgetting on the previous target domain. Thus, we add the distillation loss \cite{li2017learning} that conducts knowledge distillation from the previous segmentation network:
\begin{equation}
    L_{distill}(\hat{Y}^s, y_{old}^s) = -\sum_{i=1}^{C \times H \times W}{y}_{old}^{s(i)'} \: \log  \hat{Y}^{s(i)'},
\label{eq:distill}
\end{equation}
with
\begin{equation}
    {y}_{old}^{s(i)'}=\frac{\exp({y}_{old}^{s(i)}/\:T')}{\sum_{j}\exp({y}_{old}^{s(j)}/\:T')}, \; \hat{Y}^{s(i)'}=\frac{\exp(\hat{Y}^{s(i)}/\:T')}{\sum_{j}\exp(\hat{Y}^{s(j)}/\:T')},
\label{eq:temperature}
\end{equation}
where $T'$ is the temperature which softens the weight distributions, and $y_{old}^s$ is obtained by passing the source image through the segmentation network before training on the current target domain.

Let $f(x)$ and $D(x)$ be the outputs from the segmentation network and the discriminator for input $x$, respectively. Then, the training objective of the proposed ETM framework is as follows:
\begin{equation}
\begin{aligned}
\min \; & \Big[ \lambda_{seg} \cdot L_{seg}(\hat{y}^{s+}, y^s)
+ \lambda_{adv} \cdot L_{adv}^{DHA}(\hat{z}^{s+}, \hat{z}^{t+}) + \\ 
& \lambda_{distill} \cdot L_{distill}(\hat{y}^s, y_{old}^s)  \Big],
\end{aligned}
\label{eq:ETM1}
\end{equation}
\begin{equation}
\begin{multlined}
\min \; \Big[L_D^{DHA}(\hat{z}^{s+}, \hat{z}^{t+})\Big],
\end{multlined}
\label{eq:ETM2}
\end{equation}
with
\begin{equation}
\begin{aligned}
& \hat{y}^{s+} = f^+(x^s), \hat{y}^{s} = f(x^s), \\ 
& \hat{z}^{s+} = D\big(f^{[:k]+}(x^s)), \hat{z}^{t+} = D\big(f^{[:k]+}(x^t)),
\end{aligned}
\label{eq:ETM3}
\end{equation}
where $L_{seg}$ is the cross-entropy loss. $\lambda_{seg}$, $\lambda_{adv}$ and $\lambda_{distill}$ are hyperparameters which are weights for each loss term. 
Once the segmentation network and the TM are updated by Eq. (\ref{eq:ETM1}), the discriminator is updated by Eq. (\ref{eq:ETM2}), alternately.
Here, we assume that the source domain data are accessible in every stage since the purpose of UDA is to perform semantic segmentation for the target domains.

\begin{algorithm}[t!]
    \caption{ETM framework} \label{algorithm} 
    \begin{algorithmic}[1]
        \State \textbf{Input:} $\mathcal{S}$, $\{\mathcal{T}_i\}_{i=1}^{T}$, $f$, $\{TM_i\}_{i=1}^{T}$, $\{D_i\}_{i=1}^{T}$, learning rates \{$\alpha_{f}$, $\alpha_{TM}$, $\alpha_{D}$\}
        \State initialize $\phi_f$
        \For {$i=1,\ldots,T$}
            \State initialize $\phi_{TM_i}$, $\phi_{D_i}$
            \If{$i>1$} Record $y_{old}^{s}$ by passing $\mathcal{S}$ through $f$
            \EndIf
            \For {$iteration=1,\ldots,iter_{max}$}
                \State Sample batches of $(x^s, y^s)$ from $\mathcal{S}$ 
                \State Sample batches of $x^t$ from $\mathcal{T}_i$
                \State \begin{varwidth}[t]{\linewidth} $\phi_f \leftarrow \phi_f - \alpha_f \frac{\partial(L_{seg}+ L_{adv}^{DHA} + L_{distill})}{\partial \phi_f}$, \par 
                where $L_{distill} = 0$ for $i = 1$ (Refer to Eq. (\ref{eq:ETM1}))
                
                \end{varwidth}
                
                \State \begin{varwidth}[t]{\linewidth} $\phi_{TM_i} \leftarrow \phi_{TM_i} - \alpha_{TM} \frac{\partial(L_{seg}+L_{adv}^{DHA})}{\partial \phi_{TM_i}}$ \par
                (Refer to Eq. (\ref{eq:ETM1}))
                \end{varwidth}
                \State \begin{varwidth}[t]{\linewidth} $\phi_{D_i} \leftarrow \phi_{D_i} - \alpha_{D} \frac{\partial(L_{D}^{DHA})}{\partial \phi_{D_i}}$ (Refer to Eq. (\ref{eq:ETM2}))
                \end{varwidth}
            \EndFor
            \State \textbf{end for}
            \State Store $TM_i$ (while $D_i$ is not stored)
        \EndFor
        \State \textbf{end for} 
    \end{algorithmic} 
\end{algorithm}
\begin{table}[t!]
\footnotesize
\begin{tabular}{l}
*\:$\lambda_{seg}$, $\lambda_{adv}$, and $\lambda_{distill}$ are omitted for clarity.
\end{tabular}
\end{table}

Algorithm \ref{algorithm} summarizes the learning procedure of our framework. Here, $\phi_f$, $\phi_{TM_i}$, and $\phi_{D_i}$ indicate the parameters of the segmentation network, the $i$-th TM, and the $i$-th discriminator, respectively. In short, by applying the proposed ETM framework to the existing UDA methods, the segmentation network and all TMs learn knowledge about all the sequential target domains while effectively keeping the knowledge about the previous target domains.

\section{Experiments}
\label{sec:exp}
\subsection{Experimental Settings}
\label{sec:expsettings}
We used two synthetic road image datasets (GTA5, SYNTHIA) as the source domains and three real-world road image datasets (CityScapes, IDD, Cross-City) as the target domains.
We used three types of baseline models: \textit{(1) Source Only}, which indicates the model trained on the source domain data and tested on the target domains, \textit{(2) the UDA methods} containing FCN-W \cite{hoffman2016fcns}, AdaptSegNet \cite{tsai2018learning}, AdvEnt \cite{vu2019advent}, SIM \cite{wang2020differential}, and \textit{(3) the Continual UDA method}, ACE \cite{wu2019ace}. For a fair comparison, DeepLab-V2 \cite{chen2017deeplab}, which is one of the representative semantic segmentation networks, was used as the segmentation network in all methods.
For \textit{Ours}, we applied the ETM framework on AdaptSegNet \cite{tsai2018learning}. The evaluation metric is mIoU (mean Intersection over Union), i.e., the average of the IoU values for each object. We implemented our experiments using Pytorch of version 1.4.0 on a Ubuntu 16.04 workstation. Single RTX 2080ti GPU was used with CUDA version 10.0.

\subsection{Comparison with State-of-the-art Methods}
\label{sec:exp_sota}
To validate the effectiveness of the proposed ETM framework, we compared the model trained with our framework with the state-of-the-art methods.
In our experiments, two Continual UDA scenarios were considered: a two-target scenario and a four-target scenario. Following the prior works on UDA \cite{hoffman2016fcns,tsai2018learning,vu2019advent,zhang2019category}, the synthetic datasets, the GTA5 and SYNTHIA datasets, were used as the source domains. In the first experiment, we first performed UDA to the CityScapes dataset, followed by Continual UDA to the IDD dataset. The CityScapes dataset is the most widely used real-world image dataset in the UDA field. To maximize the catastrophic forgetting problem for the CityScapes dataset collected in European cities, we then performed UDA to the IDD dataset collected in entirely different Indian cities. In the second experiment, we performed Continual UDA sequentially for four different cities in the Cross-City dataset (Rio, Rome, Taipei, and Tokyo). Through a longer target domain sequence, we analyzed the catastrophic forgetting problem in depth.

We experimented with two different input sizes. If the original images were used as the inputs by converting into high-definition images of 1024 $\times$ 512, we specified them as H, and L if the images were converted into low-definition images of 512 $\times$ 256. We conducted the experiments using both input sizes for all methods except ACE \cite{wu2019ace}. For the ACE, experiments on high-quality images (H) were not possible due to a problem with the network capacity.

\begin{table*}[t!]
\centering
\scriptsize
\caption{\textbf{The two-target scenario results.} The best performance is presented in bold, and the second-best performance is underlined. Note that all the numbers reported represent the performance after the final domain adaptation.}
\label{table:main1}
\begin{tabular}{
@{}cc||>{\centering\arraybackslash}p{1.2cm}>{\centering\arraybackslash}p{1.2cm}>{\centering\arraybackslash}p{1.2cm}>{\centering\arraybackslash}p{1.2cm}||>{\centering\arraybackslash}p{1.2cm}>{\centering\arraybackslash}p{1.2cm}>{\centering\arraybackslash}p{1.2cm}>{\centering\arraybackslash}p{1.2cm}>{\centering\arraybackslash}p{1.2cm}>{\centering\arraybackslash}p{1.2cm}@{}}

\toprule
&  
& \multicolumn{4}{c||}{\textbf{GTA5 $\rightarrow$ CityScapes $\rightarrow$ IDD}} 
& \multicolumn{4}{c}{\textbf{SYNTHIA $\rightarrow$ CityScapes $\rightarrow$ IDD}} \\ \cmidrule(l){3-10} 
\multirow{2}{*}{\textbf{Model}} & {\begin{tabular}[c]{@{}c@{}}\textbf{Input}\\\textbf{Size}\end{tabular}} & \begin{tabular}[c]{@{}c@{}}\textbf{CityScapes}\\ \textbf{(Fgt.)}\end{tabular} & 
\textbf{IDD} & \cellcolor{gray!25}\begin{tabular}[c]{@{}c@{}}\textbf{Mean}\\ \textbf{mIoU}\end{tabular} & 
\cellcolor{gray!25}\textbf{Gain} & 
\begin{tabular}[c]{@{}c@{}}\textbf{CityScapes}\\ \textbf{(Fgt.)}\end{tabular} & 
\textbf{IDD} &
\cellcolor{gray!25}\begin{tabular}[c]{@{}c@{}}\textbf{Mean}\\ \textbf{mIoU}\end{tabular} & 
\cellcolor{gray!25}\textbf{Gain} \\ \midrule

{\textbf{Source Only}} 
& L 
& 27.55 & 37.83 & \cellcolor{gray!25}32.69 & \cellcolor{gray!25}- 
& 32.30 & 29.65 & \cellcolor{gray!25}30.98 & \cellcolor{gray!25}- \\ \midrule

\textbf{FCN-W \cite{hoffman2016fcns}} 
& L 
& \begin{tabular}[c]{@{}c@{}} 24.60\\ \tiny{(-5.71)}\end{tabular} 
& 30.50 & \cellcolor{gray!25} 27.55 & \cellcolor{gray!25} -5.14 
& \begin{tabular}[c]{@{}c@{}} 30.02\\ \tiny{(+0.71)}\end{tabular} 
& 23.95 & \cellcolor{gray!25} 26.99 & \cellcolor{gray!25} -3.99\\ \\[-0.9em]

\textbf{AdaptSegNet \cite{tsai2018learning}}
& L 
& \begin{tabular}[c]{@{}c@{}} 32.23\\ \tiny{(-2.42)}\end{tabular} 
& 41.51 & \cellcolor{gray!25}36.87 & \cellcolor{gray!25}+4.18  
& \begin{tabular}[c]{@{}c@{}} 40.10\\ \tiny{(-0.08)}\end{tabular}
& 34.79 & \cellcolor{gray!25} 37.45 & \cellcolor{gray!25}+6.47 \\ \\[-0.9em] 

\textbf{AdvEnt \cite{vu2019advent}}
& L 
& \begin{tabular}[c]{@{}c@{}} \underline{33.25}\\ \tiny{(-1.71)}\end{tabular} 
& \underline{41.60} & \cellcolor{gray!25}\underline{37.43} & \cellcolor{gray!25}\underline{+4.74}  
& \begin{tabular}[c]{@{}c@{}} \underline{40.26}\\ \tiny{(-0.26)}\end{tabular} 
& \underline{34.80} & \cellcolor{gray!25} \underline{37.53}& \cellcolor{gray!25}\underline{+6.55} \\ \\[-0.9em] 

\textbf{ACE \cite{wu2019ace}}         
& L 
& \begin{tabular}[c]{@{}c@{}}28.39\\ \tiny{(-1.82)}\end{tabular}  
& 34.70 & \cellcolor{gray!25}31.55 & \cellcolor{gray!25}-1.14  
& \begin{tabular}[c]{@{}c@{}}28.67 \\ \tiny{(-1.14)}\end{tabular} 
& 29.78 & \cellcolor{gray!25}29.23 & \cellcolor{gray!25}-1.75 \\ \\[-0.9em] 

\textbf{SIM \cite{wang2020differential}}
& L
& \begin{tabular}[c]{@{}c@{}}24.99\\ \tiny{\textbf{(-0.71)}}\end{tabular}  
& 24.99 & \cellcolor{gray!25} 24.99& \cellcolor{gray!25} -7.70  
& \begin{tabular}[c]{@{}c@{}} 31.01\\ \tiny{\underline{(-0.04)}}\end{tabular} 
& 27.77 & \cellcolor{gray!25} 29.39& \cellcolor{gray!25} -1.59\\ \midrule

\textbf{ETM (Ours)}
& L
& \begin{tabular}[c]{@{}c@{}} \textbf{34.36}\\ \tiny{\underline{(-1.26)}}\end{tabular}  
& \textbf{41.67}  & \cellcolor{gray!25}\textbf{38.02} & \cellcolor{gray!25}\textbf{+5.33} 
& \begin{tabular}[c]{@{}c@{}} \textbf{40.48}\\ \tiny{\textbf{(+0.75)}}\end{tabular} 
& \textbf{34.85} & \cellcolor{gray!25} \textbf{37.67} & \cellcolor{gray!25}\textbf{+6.69}  
\\ \cmidrule{1-10}\morecmidrules\cmidrule{1-10}

{\textbf{Source Only}} 
& H  
& 34.70 & 42.65 & \cellcolor{gray!25}38.68 & \cellcolor{gray!25}- 
& 35.02 & 31.74 & \cellcolor{gray!25}33.38 & \cellcolor{gray!25}- \\ \midrule

\textbf{FCN-W \cite{hoffman2016fcns}} 
& H 
& \begin{tabular}[c]{@{}c@{}}32.03 \\ \tiny{(-3.81)}\end{tabular} 
& 35.57 & \cellcolor{gray!25}33.80 & \cellcolor{gray!25}-4.88  
& \begin{tabular}[c]{@{}c@{}}31.12 \\ \tiny{(-2.27)} \end{tabular}  
& 29.48 & \cellcolor{gray!25}30.30 & \cellcolor{gray!25}-3.08 \\  \\[-0.9em]

\textbf{AdaptSegNet \cite{tsai2018learning}}
& H 
& \begin{tabular}[c]{@{}c@{}}34.86\\ \tiny(-7.41)\end{tabular}   
& \underline{43.87} & \cellcolor{gray!25}39.37 & \cellcolor{gray!25}+0.69  
& \begin{tabular}[c]{@{}c@{}}\underline{42.08} \\ \tiny(-3.33)\end{tabular}   
& \underline{35.64} & \cellcolor{gray!25}\underline{38.86} & \cellcolor{gray!25}\underline{+5.48} \\ \\[-0.9em]

\textbf{AdvEnt \cite{vu2019advent}}
& H 
& \begin{tabular}[c]{@{}c@{}}\underline{37.27} \\ \tiny(-5.86)\end{tabular}   
& 43.41 & \cellcolor{gray!25}\underline{40.34} & \cellcolor{gray!25}\underline{+1.66}  
& \begin{tabular}[c]{@{}c@{}}39.14 \\ \tiny(-6.39) \end{tabular}  
& 32.15 & \cellcolor{gray!25}35.65 & \cellcolor{gray!25}+2.27 \\ \\[-0.9em]

\textbf{SIM \cite{wang2020differential}}         
& H 
& \begin{tabular}[c]{@{}c@{}}35.47\\ \tiny{\textbf{(-0.64)}}\end{tabular}  
& 33.93 & \cellcolor{gray!25} 34.70& \cellcolor{gray!25} -3.98
& \begin{tabular}[c]{@{}c@{}} 39.19\\ \tiny{\textbf{(-1.35)}}\end{tabular} 
& 32.77 & \cellcolor{gray!25} 35.98& \cellcolor{gray!25} +2.60\\ \midrule

\textbf{ETM (Ours)}
& H 
& \begin{tabular}[c]{@{}c@{}} \textbf{40.61}\\ \tiny{\underline{(-1.35)}}\end{tabular}   
& \textbf{46.73}  & \cellcolor{gray!25}\textbf{43.67} & \cellcolor{gray!25}\textbf{+4.99} 
& \begin{tabular}[c]{@{}c@{}} \textbf{43.86}\\\tiny{\underline{(-1.92)}}\end{tabular}   
& \textbf{37.17} & \cellcolor{gray!25}\textbf{40.52} & \cellcolor{gray!25}\textbf{+7.14}                                     \\ \bottomrule

\end{tabular}
\end{table*}

\paragraph{Two-Target Scenario.}

The results of Continual UDA on two target domains (CityScapes and IDD), using each of the GTA5 and SYNTHIA datasets as a source domains are shown in Table \ref{table:main1}. 
\textit{Mean mIoU} is the average of the mIoU values for the two target domains. In the \textit{Gain} column, we specify a difference in the mean mIoU values compared to Source Only. In the CityScapes column, forgetting values (Fgt.) along with the mIoU values are also shown in parentheses. The forgetting value is the difference between the mIoU values for the CityScapes dataset when the network is adapted to the CityScapes dataset and when the network is continually adapted to the IDD dataset. The larger the number in parentheses, the less the forgetting on the previous domain.

From the results of Table \ref{table:main1}, when the ETM framework is applied, we can see that in most cases, not only the semantic segmentation performance is the best, but also the least forgetting occurs. Since the TM is proposed to overcome the catastrophic forgetting problem in Continual UDA, the low forgetting on the CityScapes dataset demonstrates that the design of the TM is valid. Furthermore, when our framework is applied, the performance for the both target domains is also high. This fact allows us to confirm that the UDA performance itself is increased by using the DHA loss. It is considered that the segmentation network learns domain-invariant information efficiently while the TM learns the knowledge corresponding to the target domain shift. 

The ACE is a method designed to adapt to changing environments such as altering weather and lighting conditions. However, experimental results of the ACE show severe forgetting and the low performance. This indicates that the style transfer method is unsuitable for the real-world problem with large domain discrepancy, and demonstrates that our framework can handle such problem.
Note that within a limited capacity network, there is a \textit{trade-off} between adapting to the current target domain and maintaining the knowledge of the previous target domain. Regarding the SIM results, although numerically less forgetting occurs, it is futile since the average adaptation performance is remarkably deficient. On the other hand, the model with our framework overcomes such \textit{trade-off}.

\begin{table*}[t!]
\centering
\scriptsize
\caption{\textbf{The four-target scenario results.} The best performance is presented in bold, and the second-best performance is underlined. Note that all the numbers reported represent the performance after the final domain adaptation.}
\label{table:main2}
\begin{tabular}{@{}cc||>{\centering\arraybackslash}p{0.75cm}>{\centering\arraybackslash}p{0.75cm}>{\centering\arraybackslash}p{0.75cm}>{\centering\arraybackslash}p{0.75cm}>{\centering\arraybackslash}p{0.75cm}>{\centering\arraybackslash}p{0.75cm}||>{\centering\arraybackslash}p{0.75cm}>{\centering\arraybackslash}p{0.75cm}>{\centering\arraybackslash}p{0.75cm}>{\centering\arraybackslash}p{0.75cm}>{\centering\arraybackslash}p{0.75cm}>{\centering\arraybackslash}p{0.75cm}>{\centering\arraybackslash}p{0.75cm}>{\centering\arraybackslash}p{0.75cm}@{}}

\toprule
&  
& \multicolumn{6}{c||}{\textbf{GTA5 $\rightarrow$ Rio $\rightarrow$ Rome $\rightarrow$ Taipei $\rightarrow$ Tokyo}} 
& \multicolumn{6}{c}{\textbf{SYNTHIA $\rightarrow$ Rio $\rightarrow$ Rome $\rightarrow$ Taipei $\rightarrow$ Tokyo}} \\ \cmidrule(l){3-14} 
\multirow{2}{*}{\textbf{Model}} & {\begin{tabular}[c]{@{}c@{}}\textbf{Input}\\\textbf{Size}\end{tabular}} & \begin{tabular}[c]{@{}c@{}}\textbf{Rio}\\ \textbf{(Fgt.)}\end{tabular}  &
\begin{tabular}[c]{@{}c@{}}\textbf{Rome}\\ \textbf{(Fgt.)}\end{tabular}  &
\begin{tabular}[c]{@{}c@{}}\textbf{Taipei}\\ \textbf{(Fgt.)}\end{tabular}  &
\textbf{Tokyo} & \cellcolor{gray!25}\begin{tabular}[c]{@{}c@{}}\textbf{Mean}\\ \textbf{mIoU}\end{tabular} & \cellcolor{gray!25}\textbf{Gain} & 
\begin{tabular}[c]{@{}c@{}}\textbf{Rio}\\ \textbf{(Fgt.)}\end{tabular}  &
\begin{tabular}[c]{@{}c@{}}\textbf{Rome}\\ \textbf{(Fgt.)}\end{tabular}  &
\begin{tabular}[c]{@{}c@{}}\textbf{Taipei}\\ \textbf{(Fgt.)}\end{tabular}  &
\textbf{Tokyo} & \cellcolor{gray!25}\begin{tabular}[c]{@{}c@{}}\textbf{Mean}\\ \textbf{mIoU}\end{tabular} & \cellcolor{gray!25}\textbf{Gain} \\ \midrule

{\textbf{Source Only}} 
& L 
& 36.98 & 38.05 & 33.94 & 33.85 & \cellcolor{gray!25}35.71 & \cellcolor{gray!25}- 
& 33.50 & 31.30 & 30.17 & 29.44 & \cellcolor{gray!25}31.10 & \cellcolor{gray!25}- \\ \midrule

\textbf{FCN-W \cite{hoffman2016fcns}}  
& L
& \begin{tabular}[c]{@{}c@{}} 24.98\\ \tiny{(-1.96)}\end{tabular} 
& \begin{tabular}[c]{@{}c@{}} 25.56\\ \tiny{(-2.90)}\end{tabular}
& \begin{tabular}[c]{@{}c@{}} 24.63\\ \tiny{(-1.33)}\end{tabular} 
& 29.68 & \cellcolor{gray!25}26.21 & \cellcolor{gray!25}-9.50  
& \begin{tabular}[c]{@{}c@{}} 25.12\\ \tiny{(-1.83)}\end{tabular} 
& \begin{tabular}[c]{@{}c@{}} 27.01\\ \tiny{\underline{(+0.12)}}\end{tabular} 
& \begin{tabular}[c]{@{}c@{}} 23.84\\ \tiny{(-0.08)}\end{tabular} 
& 26.84 & \cellcolor{gray!25}25.70 & \cellcolor{gray!25}-5.40 \\ \\[-0.9em]

\textbf{AdaptSegNet \cite{tsai2018learning}}
& L 
& \begin{tabular}[c]{@{}c@{}} \underline{38.68}\\ \tiny{(-1.81)}\end{tabular} 
& \begin{tabular}[c]{@{}c@{}} \underline{39.99}\\ \tiny{\underline{(-0.18)}}\end{tabular} 
& \begin{tabular}[c]{@{}c@{}} 34.17\\ \tiny{\underline{(-0.06)}}\end{tabular} 
& 36.35 & \cellcolor{gray!25} 37.30 & \cellcolor{gray!25}+1.59  
& \begin{tabular}[c]{@{}c@{}} 32.95\\ \tiny{(-2.64)}\end{tabular} 
& \begin{tabular}[c]{@{}c@{}} \underline{32.02}\\ \tiny{(-0.08)}\end{tabular}
& \begin{tabular}[c]{@{}c@{}} 29.30\\ \tiny{(-0.38)}\end{tabular} 
& 28.36 & \cellcolor{gray!25} 30.66 & \cellcolor{gray!25}-0.44 \\ \\[-0.9em] 

\textbf{AdvEnt \cite{vu2019advent}}
& L 
& \begin{tabular}[c]{@{}c@{}} 37.88\\ \tiny{(-3.53)}\end{tabular} 
& \begin{tabular}[c]{@{}c@{}} 39.64\\ \tiny{(-1.64)}\end{tabular} 
& \begin{tabular}[c]{@{}c@{}} \underline{36.06}\\ \tiny{(-1.01)}\end{tabular} 
& \underline{37.84} & \cellcolor{gray!25} \underline{37.86} & \cellcolor{gray!25}\underline{+2.15}  
& \begin{tabular}[c]{@{}c@{}} \underline{32.97}\\ \tiny{(-2.50)}\end{tabular} 
& \begin{tabular}[c]{@{}c@{}} 31.50\\ \tiny{(+0.11)}\end{tabular} 
& \begin{tabular}[c]{@{}c@{}} \underline{29.37}\\ \tiny{\underline{(-0.04)}}\end{tabular} 
& 28.82 & \cellcolor{gray!25} \underline{30.67} & \cellcolor{gray!25}\underline{-0.43}\\ \\[-0.9em] 

\textbf{ACE \cite{wu2019ace}}          
& L 
&  \begin{tabular}[c]{@{}c@{}} 30.78\\ \tiny{(-4.70)}\end{tabular} 
&  \begin{tabular}[c]{@{}c@{}} 31.35\\ \tiny{(-1.12)}\end{tabular} 
&  \begin{tabular}[c]{@{}c@{}} 28.56\\ \tiny{(-1.16)}\end{tabular} 
& 33.53 & \cellcolor{gray!25}31.06 & \cellcolor{gray!25}-4.65  
&  \begin{tabular}[c]{@{}c@{}} 27.27\\ \tiny{(-6.53)}\end{tabular} 
&  \begin{tabular}[c]{@{}c@{}} 28.26\\ \tiny{(-2.69)}\end{tabular} 
&  \begin{tabular}[c]{@{}c@{}} 26.39\\ \tiny{(-0.43)}\end{tabular} 
& \underline{29.05} & \cellcolor{gray!25}27.74 & \cellcolor{gray!25}-3.36 \\ \\[-0.9em] 

\textbf{SIM \cite{wang2020differential}}           
& L 
&  \begin{tabular}[c]{@{}c@{}} 27.62\\ \tiny{\underline{(-1.23)}}\end{tabular} 
&  \begin{tabular}[c]{@{}c@{}} 27.48\\ \tiny{(-0.54)}\end{tabular} 
&  \begin{tabular}[c]{@{}c@{}} 24.38\\ \tiny{(-0.07)}\end{tabular} 
& 28.92 & \cellcolor{gray!25} 27.10 & \cellcolor{gray!25}-8.61 
&  \begin{tabular}[c]{@{}c@{}} 27.82\\ \tiny{\underline{(-1.04)}}\end{tabular} 
&  \begin{tabular}[c]{@{}c@{}} 28.08\\ \tiny{(-1.49)}\end{tabular} 
&  \begin{tabular}[c]{@{}c@{}} 26.09\\ \tiny{(-0.43)}\end{tabular} 
& 27.84 & \cellcolor{gray!25} 27.46& \cellcolor{gray!25}-3.64 \\ \midrule

\textbf{ETM (Ours)}  
& L 
& \begin{tabular}[c]{@{}c@{}} \textbf{41.15}\\ \tiny{\textbf{(+0.86)}}\end{tabular}   
& \begin{tabular}[c]{@{}c@{}} \textbf{40.76}\\ \tiny{\textbf{(+0.31)}}\end{tabular} 
& \begin{tabular}[c]{@{}c@{}} \textbf{37.12}\\ \tiny{\textbf{(+0.83)}}\end{tabular}
& \textbf{37.94}  & \cellcolor{gray!25}\textbf{39.24} & \cellcolor{gray!25}\textbf{+3.53} 
& \begin{tabular}[c]{@{}c@{}} \textbf{34.97}\\ \tiny{\textbf{(+0.30)}}\end{tabular}
& \begin{tabular}[c]{@{}c@{}} \textbf{33.74}\\ \tiny{\textbf{(+1.19)}}\end{tabular}
& \begin{tabular}[c]{@{}c@{}} \textbf{30.81}\\ \tiny{\textbf{(+0.15)}}\end{tabular}
& \textbf{31.75} & \cellcolor{gray!25}\textbf{32.82} & \cellcolor{gray!25}\textbf{+1.72}
\\ \cmidrule{1-14}\morecmidrules\cmidrule{1-14}

{\textbf{Source Only}} 
& H 
& 44.21 & 44.41 & 40.79 & 43.99 & \cellcolor{gray!25}43.35 & \cellcolor{gray!25}- 
& 36.63 & 32.11 & 31.33 & 31.74 & \cellcolor{gray!25}32.95 & \cellcolor{gray!25}- \\ \midrule

\textbf{FCN-W \cite{hoffman2016fcns}} 
& H 
&  \begin{tabular}[c]{@{}c@{}}32.65 \\ \tiny{(-5.65)}\end{tabular} 
&  \begin{tabular}[c]{@{}c@{}}32.59 \\ \tiny{(-2.01)}\end{tabular} 
&  \begin{tabular}[c]{@{}c@{}}28.55 \\ \tiny{\textbf{(+0.09)}}\end{tabular} 
& 35.27 & \cellcolor{gray!25}32.27 & \cellcolor{gray!25}-11.08  
& \begin{tabular}[c]{@{}c@{}}30.45  \\ \tiny{(-4.79)}\end{tabular} 
& \begin{tabular}[c]{@{}c@{}}30.35  \\ \tiny{\underline{(-0.34)}}\end{tabular} 
& \begin{tabular}[c]{@{}c@{}}26.42  \\ \tiny{(-1.55)}\end{tabular} 
& \underline{31.76} & \cellcolor{gray!25}29.75 & \cellcolor{gray!25}-3.20 \\ \\[-0.9em] 

\textbf{AdaptSegNet \cite{tsai2018learning}}
& H 
& \begin{tabular}[c]{@{}c@{}} 43.00 \\ \tiny{(-8.32)}\end{tabular} 
& \begin{tabular}[c]{@{}c@{}} 43.54 \\ \tiny{(-3.68)}\end{tabular} 
& \begin{tabular}[c]{@{}c@{}} 36.94 \\ \tiny{(-2.86)}\end{tabular} 
& 42.38 & \cellcolor{gray!25}41.47 & \cellcolor{gray!25}-1.88  
& \begin{tabular}[c]{@{}c@{}} 36.54 \\ \tiny{(-2.56)}\end{tabular} 
& \begin{tabular}[c]{@{}c@{}} 33.46 \\ \tiny{(-1.69)}\end{tabular} 
& \begin{tabular}[c]{@{}c@{}} 31.45 \\ \tiny{(-0.44)}\end{tabular} 
& 31.15 & \cellcolor{gray!25}33.15 & \cellcolor{gray!25}+0.20 \\ \\[-0.9em]

\textbf{AdvEnt \cite{vu2019advent}}
& H 
& \begin{tabular}[c]{@{}c@{}} \underline{43.60} \\ \tiny{(-5.93)}\end{tabular} 
& \begin{tabular}[c]{@{}c@{}} \underline{45.08} \\ \tiny{(-2.78)}\end{tabular} 
& \begin{tabular}[c]{@{}c@{}} \underline{37.55} \\ \tiny{(-1.72)}\end{tabular} 
& \underline{44.12} & \cellcolor{gray!25}\underline{42.59} & \cellcolor{gray!25}\underline{-0.76}  
& \begin{tabular}[c]{@{}c@{}} \underline{37.66} \\ \tiny{(-2.04)}\end{tabular} 
& \begin{tabular}[c]{@{}c@{}} \underline{34.41} \\ \tiny{(-1.47)}\end{tabular} 
& \begin{tabular}[c]{@{}c@{}} \underline{32.32} \\ \tiny{\underline{(-0.25)}}\end{tabular} 
& 31.35 & \cellcolor{gray!25}\underline{33.94} & \cellcolor{gray!25}\underline{+0.99} \\ \\[-0.9em]

\textbf{SIM \cite{wang2020differential}}           
& H 
&  \begin{tabular}[c]{@{}c@{}} 39.65\\ \tiny{\underline{(-1.86)}}\end{tabular} 
&  \begin{tabular}[c]{@{}c@{}} 37.61\\ \tiny{\underline{(-1.71)}}\end{tabular} 
&  \begin{tabular}[c]{@{}c@{}} 35.72\\ \tiny{(-1.19)}\end{tabular} 
& 38.24 & \cellcolor{gray!25}37.81 & \cellcolor{gray!25} -5.54
&  \begin{tabular}[c]{@{}c@{}} 34.30\\ \tiny{\underline{(-1.02)}}\end{tabular} 
&  \begin{tabular}[c]{@{}c@{}} 32.82\\ \tiny{(-0.68)}\end{tabular} 
&  \begin{tabular}[c]{@{}c@{}} 31.48\\ \tiny{(-0.53)}\end{tabular} 
& 30.73 & \cellcolor{gray!25} 32.33& \cellcolor{gray!25}-0.62 \\ \midrule

\textbf{ETM (Ours)}               
& H 
& \begin{tabular}[c]{@{}c@{}} \textbf{48.45}\\ \tiny{\textbf{(-1.79)}}\end{tabular}  
& \begin{tabular}[c]{@{}c@{}} \textbf{45.93}\\ \tiny{\textbf{(-0.38)}}\end{tabular} 
& \begin{tabular}[c]{@{}c@{}} \textbf{41.19}\\ \tiny{\underline{(-0.07)}}\end{tabular} 
& \textbf{44.18}  & \cellcolor{gray!25}\textbf{44.94} & \cellcolor{gray!25}\textbf{+1.59} 
& \begin{tabular}[c]{@{}c@{}} \textbf{38.44}\\ \tiny{\textbf{(+0.16)}}\end{tabular} 
& \begin{tabular}[c]{@{}c@{}} \textbf{35.73}\\ \tiny{\textbf{(+0.01)}}\end{tabular} 
& \begin{tabular}[c]{@{}c@{}} \textbf{32.52}\\ \tiny{\textbf{(+0.07)}}\end{tabular} 
& \textbf{34.12} & \cellcolor{gray!25}\textbf{35.20} & \cellcolor{gray!25}\textbf{+2.25}                                     \\ \bottomrule

\end{tabular}
\end{table*}

\paragraph{Four-Target Scenario.}

We conducted experiments on more target domains, since considering two target domains is not enough to deal with the catastrophic forgetting problem in Continual UDA. The results of Continual UDA for Rio, Rome, Taipei, and Tokyo in the Cross-City dataset are shown in Table \ref{table:main2}. The forgetting value in the parenthesis denotes the difference between the performances on particular target domain data when UDA has been performed to the last target domain (Tokyo) and when UDA has been performed to the particular target domain data. For example, forgetting for Taipei indicates the difference between the performances on Taipei when UDA has been done to Tokyo and when UDA has been done to Taipei.
As can be seen from Table \ref{table:main2}, the model trained with the ETM framework shows the highest semantic segmentation performance and the least forgetting in most cases. 

\begin{table}[t]
\centering
\tiny
\caption{Ablation study on each component of the ETM framework.}
\label{table:TM,DHAablation}

\begin{tabular}{@{}c||ccc||ccc@{}}
\toprule
\multirow{4}{*}{\textbf{Model}}
& \multicolumn{3}{c||}{\textbf{GTA5 $\rightarrow$ CityScapes $\rightarrow$ IDD}} 
& \multicolumn{3}{c}{\textbf{SYNTHIA $\rightarrow$ CityScapes $\rightarrow$ IDD}} \\ \cmidrule(l){2-7} 
& \begin{tabular}[c]{@{}c@{}}\textbf{CityScapes}\\ \textbf{(Fgt.)}\end{tabular} & \textbf{IDD} 
& \begin{tabular}[c]{@{}c@{}}\textbf{Mean}\\ \textbf{mIoU}\end{tabular} 
& \begin{tabular}[c]{@{}c@{}}\textbf{CityScapes}\\ \textbf{(Fgt.)}\end{tabular} & \textbf{IDD} 
& \begin{tabular}[c]{@{}c@{}}\textbf{Mean}\\ \textbf{mIoU}\end{tabular} \\ \midrule

\multicolumn{1}{l||}{\textbf{AdaptSegNet \cite{tsai2018learning}}}
& \begin{tabular}[c]{@{}c@{}} 34.86 \\ \tiny{(-7.41)}\end{tabular} 
& 43.87    & 39.37              
& \begin{tabular}[c]{@{}c@{}} 42.08 \\ \tiny{(-3.33)}\end{tabular}        
& 35.64    & 38.86  \\ \midrule

\multicolumn{1}{l||}{\begin{tabular}[l]{@{}l@{}}\textbf{AdaptSegNet \cite{tsai2018learning}}\\ \textbf{+ TM}\end{tabular}}
& \begin{tabular}[c]{@{}c@{}} \underline{40.04} \\ \tiny{\underline{(-1.45)}}\end{tabular} 
& \underline{44.28}    & \underline{42.16}                   
& \begin{tabular}[c]{@{}c@{}} \underline{43.56} \\ \tiny{\textbf{(-1.54)}}\end{tabular} 
& \underline{36.52}    & \underline{40.04}  \\ \midrule

\multicolumn{1}{l||}{\begin{tabular}[l]{@{}l@{}}\textbf{AdaptSegNet \cite{tsai2018learning}}\\ \textbf{+ TM + DHA}\end{tabular}}     
& \begin{tabular}[c]{@{}c@{}} \textbf{40.61} \\ \tiny{\textbf{(-1.35)}}\end{tabular} 
& \textbf{46.73}    & \textbf{43.67}         
& \begin{tabular}[c]{@{}c@{}} \textbf{43.86} \\ \tiny{\underline{(-1.92)}}\end{tabular} 
& \textbf{37.17}    & \textbf{40.52} \\ \bottomrule
\end{tabular}
\end{table}

\begin{table}[t!]
\centering
\tiny
\caption{Ablation study on each module of the TM. }
\label{table:TMdesign}

\begin{tabular}{@{}cc||ccc||ccc@{}}
\toprule
\multicolumn{2}{c||}{\textbf{TM Architecture}} 
& \multicolumn{3}{c||}{\textbf{GTA5 $\rightarrow$ CityScapes $\rightarrow$ IDD}} 
& \multicolumn{3}{c}{\textbf{SYNTHIA $\rightarrow$ CityScapes $\rightarrow$ IDD}} \\ \midrule
\textbf{1$\times$1 Conv.}     & \textbf{Avg. Pool.}     
& \begin{tabular}[c]{@{}c@{}}\textbf{CityScapes}\\ \textbf{(Fgt.)}\end{tabular} & \textbf{IDD} 
& \begin{tabular}[c]{@{}c@{}}\textbf{Mean}\\ \textbf{mIoU}\end{tabular} 
& \begin{tabular}[c]{@{}c@{}}\textbf{CityScapes}\\ \textbf{(Fgt.)}\end{tabular} & \textbf{IDD} 
& \begin{tabular}[c]{@{}c@{}}\textbf{Mean}\\ \textbf{mIoU}\end{tabular} \\ \midrule

\xmark       & \xmark
& \begin{tabular}[c]{@{}c@{}} 34.86 \\ \tiny{(-7.41)}\end{tabular} 
& 43.87    & 39.37              
& \begin{tabular}[c]{@{}c@{}} 42.08 \\ \tiny{(-3.33)}\end{tabular}        
& 35.64    & 38.86  \\ \midrule

\cmark       & \xmark           
& \begin{tabular}[c]{@{}c@{}} \underline{38.81} \\ \tiny{\underline{(-3.01)}}\end{tabular}
& \textbf{44.43}    & \underline{41.62}               
& \begin{tabular}[c]{@{}c@{}} \underline{42.94} \\ \tiny{\underline{(-2.17)}}\end{tabular}     
& \underline{36.00}    & \underline{39.47}  \\ \midrule

\xmark       & \cmark      
& \begin{tabular}[c]{@{}c@{}} 36.70 \\ \tiny{(-4.53)}\end{tabular}
& 44.04    & 40.37                   
& \begin{tabular}[c]{@{}c@{}} 42.22 \\ \tiny{(-2.37)}\end{tabular}
& 35.48    & 38.85  \\ \midrule

\cmark       & \cmark      
& \begin{tabular}[c]{@{}c@{}} \textbf{40.04} \\ \tiny{\textbf{(-1.45)}}\end{tabular}
& \underline{44.28}    & \textbf{42.16}         
& \begin{tabular}[c]{@{}c@{}} \textbf{43.56} \\ \tiny{\textbf{(-1.54)}}\end{tabular} 
& \textbf{36.52}    & \textbf{40.04} \\ \bottomrule
\end{tabular}
\end{table}

\subsection{Ablation Study}
\label{sec:ablation}
We conducted in-depth analyses of our framework internally by checking how the TM and the DHA loss affect the performance of the framework. All experiments in this section were conducted in an environment performing Continual UDA on the CityScapes and IDD datasets, with an input size of H.

First of all, we analyzed the effectiveness of each component of our framework. In Table \ref{table:TM,DHAablation}, the first row indicates our model without both the TM and the DHA loss since we applied our framework to the AdaptSegNet. When comparing the results in the second row and the first row, the forgetting is alleviated as well as the mIoU values for the CityScapes dataset are increased. Before adapting to the IDD dataset, the performance on the CityScapes dataset is similar to each other in both cases. That is, adding TM results in mitigation of the catastrophic forgetting problem. Moreover, when comparing the third row to the second row, the overall UDA performance is improved. It means that the UDA performance is enhanced due to the DHA loss.

\begin{figure*}[!t]
\centering  
\subfigure[]{\includegraphics[width=0.51\linewidth]{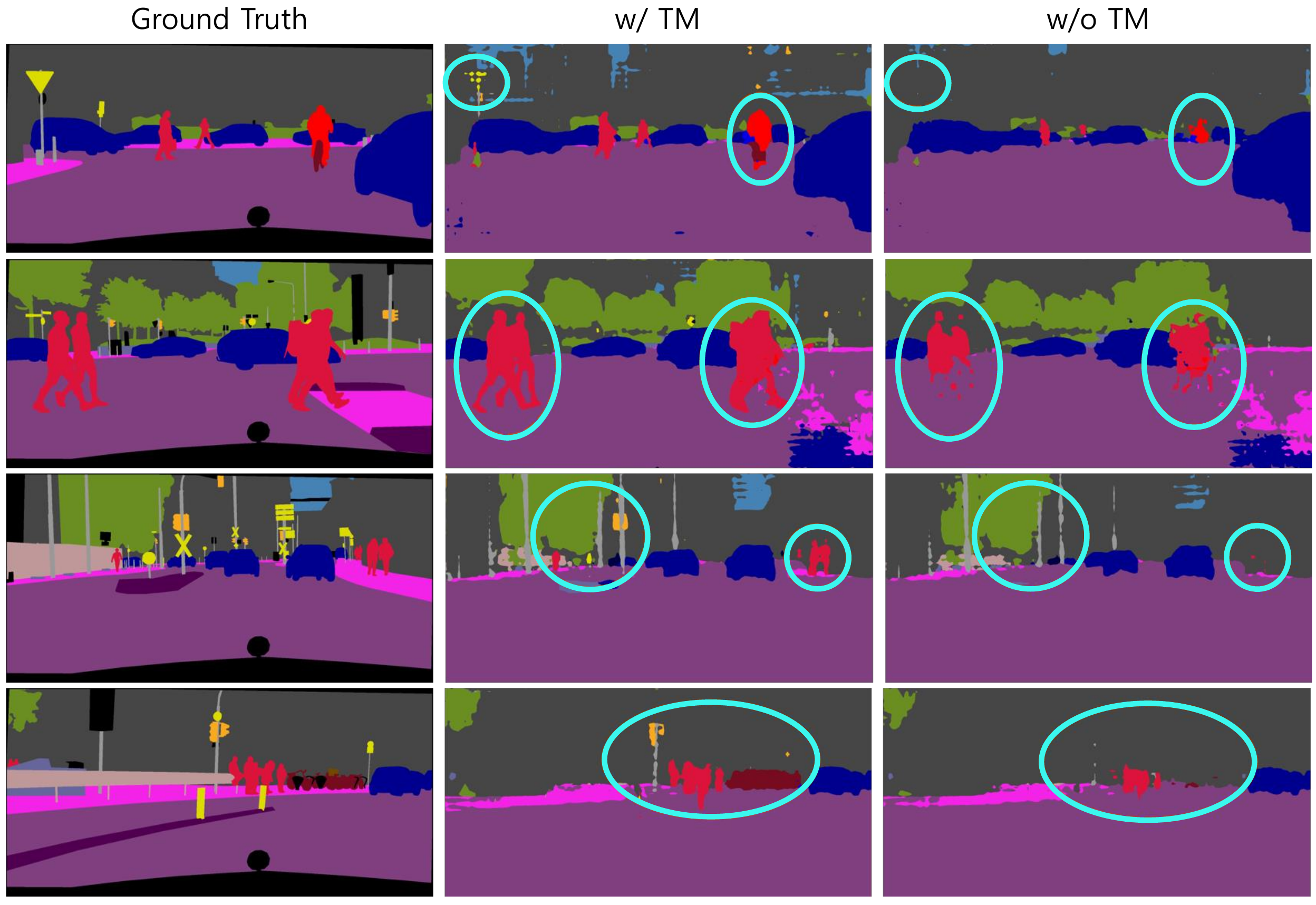}}
\subfigure[]{\includegraphics[width=0.445\linewidth]{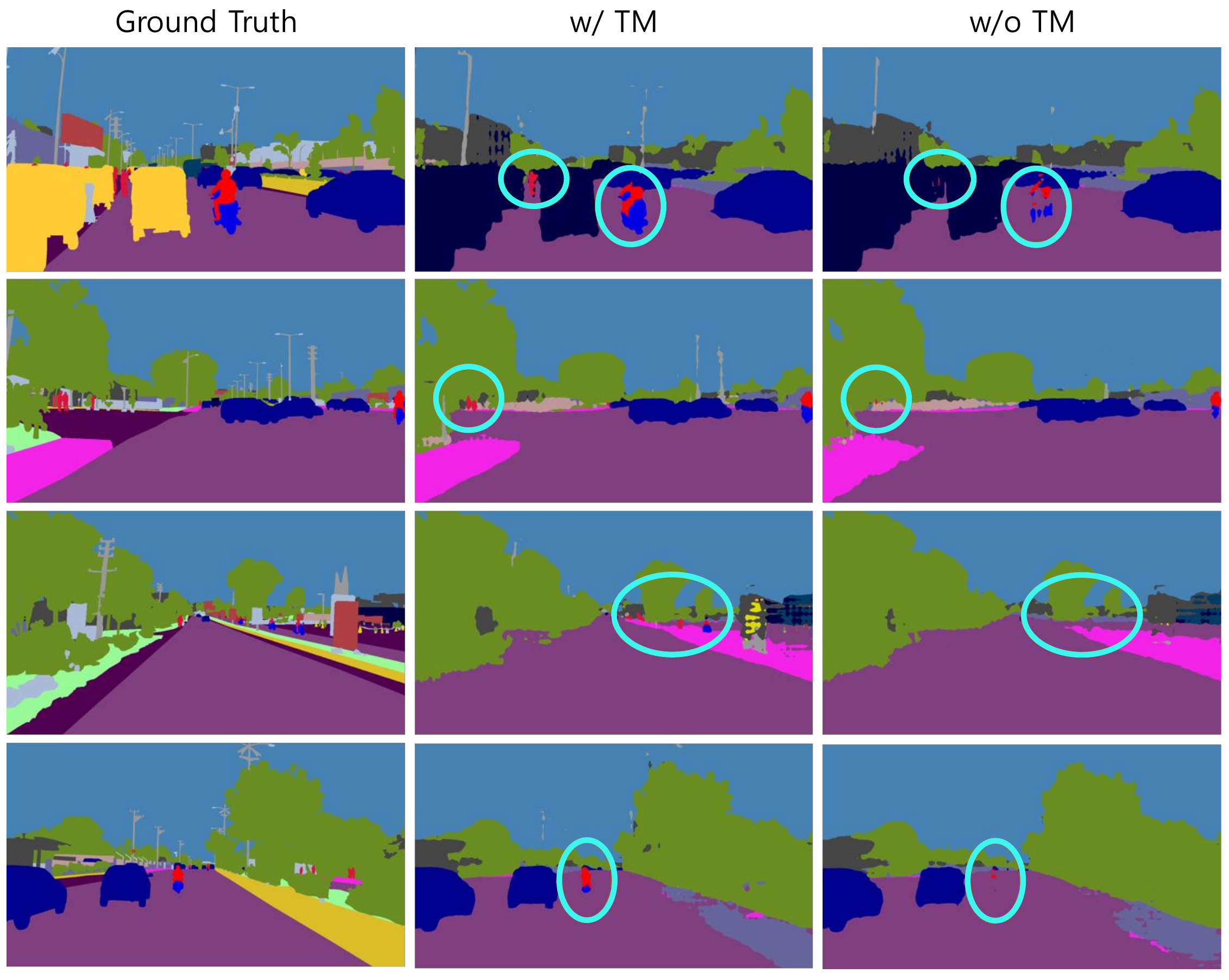}}
\caption{Comparison of the semantic maps according to the presence and absence of TM. Examples for the semantic maps of the images drawn from (a) the CityScapes dataset and (b) the IDD dataset are shown in the figure. Inside the sky-blue circles, difference between the semantic maps predicted by the network with the TM and without the TM can be observed.}
\label{fig:TMablation}
\end{figure*}

Moreover, we further analyzed how each module of the TM affects the overall performance. The differences in performance depending on the presence or absence of each module are summarized in Table \ref{table:TMdesign}. 
For a fair comparison, we set all the other conditions same except the TM architecture. To verify the effect of each module clearly, the DHA loss was not used. Firstly, note that even using one module as the TM leads to alleviating catastrophic forgetting. Also, the catastrophic forgetting problem is more mitigated when the $1\times1$ Conv. module is used compared to when the Avg. Pool. module is used. This indicates that localized information extracted by the $1\times1$ Conv. module contributes more than contextual information extracted by the Avg. Pool. module for the domain discrepancy. Finally, when both modules are used, the least forgetting occurs. From this, it can be claimed that there exists a synergy between the two modules, and the validity of the TM's design is demonstrated.

\subsection{Discussion}
\label{sec:discussion}

Previously, we claimed that the TM stores domain discrepancy information of each target domain by attaching to the segmentation network and being trained together; however, it is vague that exactly which information is stored in the TM. To clarify this, we carried out semantic segmentation using the network with the ETM framework. By comparing the results of semantic segmentation with the segmentation network only and with the integration of the segmentation network and the TM, it would be revealed which information is contained in the TM. 

Fig. \ref{fig:TMablation} shows the semantic maps predicted by the network when Continual UDA is performed to the CityScapes dataset and the IDD dataset, respectively, with the GTA5 dataset as the source domain. For each of the four images extracted from the CityScapes and IDD dataset, the ground truth of each image and the semantic maps predicted with the TM (w/ TM), and without the TM (w/o TM) are compared with each other. Fig. \ref{fig:TMablation}(a) is an example for the CityScapes dataset. In the first image of \ref{fig:TMablation}(a), a traffic sign (expressed in yellow) and a person (expressed in red) riding bicycles (expressed in brown) are well shown inside the sky-blue circle when the TM is used, but disappeared when the TM is not used. Likewise, in other images, people, traffic lights (expressed in orange), and bicycles are properly predicted with the TM, but disappeared without the TM. From Fig. \ref{fig:TMablation}(b) for the IDD dataset, we can verify that people and motorcycles (expressed in blue) are entirely missed when without the TM. 

In summary, the TM contains information about all objects in general, but especially more about the types of objects that are likely to differ from one instance to another. As stated in Section \ref{sec:TM}, the $1\times1$ Conv. module of the TM is added to extract more localized information and the Avg. Pool. module is added for broader contextual information. Through the experiment, it can be claimed that the $1\times1$ Conv. module plays a more vital role and such specific information extracted by the module contributes more in overcoming the domain discrepancy.

\section{Conclusion}
In this paper, we proposed Continual UDA for semantic segmentation based on the ETM framework. The framework can be applied to the methods performing UDA with adversarial learning. By attaching the proposed TM to the segmentation network, we alleviated the catastrophic forgetting problem that occurs when existing UDA methods are applied in continual learning environments. The TM was initiated for each target domain and effectively captured the domain discrepancy only with a small capacity. Furthermore, the proposed DHA loss enhanced the UDA performance.
By proposing the ETM framework, we have enlarged the scope of Continual UDA for semantic segmentation, and overcome the catastrophic forgetting problem in UDA. However, we hope to further develop our framework to increase the segmentation performance to the level of the model trained in a supervised manner.

{\small
\bibliographystyle{ieee_fullname}
\bibliography{ref}
}

\begin{figure*}[!t]
\centering
\includegraphics[width=0.7\linewidth]{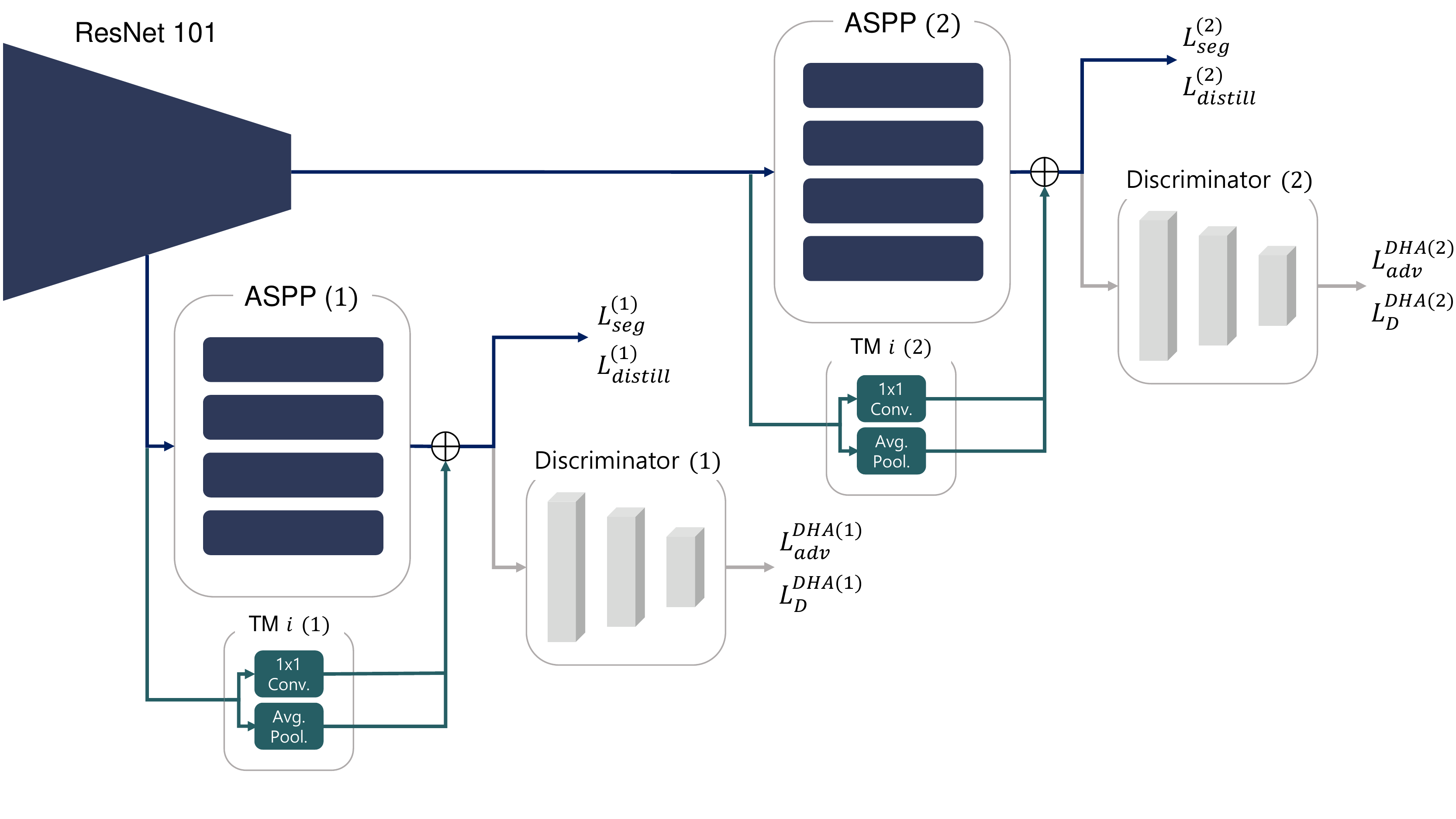}
\caption{Our ETM framework based on the AdaptSegNet. The figure describes the situation when conducting UDA for the $i$-th target domain.}
\label{fig:network}
\end{figure*}

\appendix
\addcontentsline{toc}{section}{Appendices}

\section*{Appendix}

\section{Dataset Details}
GTA5 dataset contains 24,966 images with pixel-level semantic annotations synthesized by a photorealistic open-world computer game. Also, we used SYNTHIA-RAND-CITYSCAPES (SYNTHIA) dataset which consists of 9,400 photo-realistic frames rendered from a virtual city environment. We randomly split these datasets, which were not split originally, into the training set and the validation set by the ratio of 6:1 as the same ratio as in CityScapes.

CityScapes dataset consists of 5,000 frames which are from a diverse set of stereo video sequences recorded in street scenes from 50 different cities of Europe. The dataset is split into a training, validation, and test set. Following prior works on UDA \cite{hoffman2016fcns,tsai2018learning,vu2019advent}, we evaluated the performance on the validation set of the dataset.
IDD (India Driving Dataset) has 10,003 images collected from 182 drive sequences on Indian roads. Unlike the CityScapes dataset, IDD is collected from an unstructured environment characterized as features, such as ambiguous road boundaries, muddy drivable areas, and animals' presence. The dataset is split into a training, validation, and test set. We also evaluated the performance on the validation set of the dataset following the CityScapes dataset.
Cross-City dataset consists of road images collected from four cities: Rio, Rome, Taipei, and Tokyo. The training set is not labeled, but the images in the test set are labeled.

Since the source and target domains share the same label space in the UDA task, we specify the number of objects that commonly exist across the datasets. The GTA5, CityScapes, and IDD datasets share 18 semantic classes, while the SYNTHIA, CityScapes, and IDD datasets share 13 classes. And the GTA5, SYNTHIA, Cross-City datasets have 13 classes commonly.

\section{Network Structure Details}
In our experiments, we used AdaptSegNet \cite{tsai2018learning} as the UDA method to apply our ETM framework. 
The AdaptSegNet is based on the DeepLab-V2 architecture. 
The DeepLab-V2 consists of a ResNet 101 \cite{he2016deep} module (as a feature extractor), and an Atrous Spatial Pyramid Pooling (ASPP) module (as a classifier). 
In Fig. \ref{fig:network}, the ResNet 101 and ASPP (2) indicate the feature extractor and the classifier of the DeepLab-V2, respectively.
Further, in the AdaptSegNet, ASPP (1) is added after the intermediate output of the ResNet 101 module.
The output values from the ASPP (1) and ASPP (2) modules are used as inputs to Discriminator (1) and Discriminator (2), respectively, for adversarial learning.
Since adversarial learning procedures are conducted in parallel, this is called multi-level adversarial learning. 
In the ETM framework, the TM is added right before the discriminator, thus, we added our TM to both ASPP (1) and ASPP (2) module. 
As depicted, for an arbitrary $i$-th target domain, TM $i$ (1) and TM $i$ (2) are added.
Our training objective is then expressed as follows:
\begin{equation}
\begin{aligned}
\min \; \sum_{n=1}^2 & \Big[ \lambda_{seg}^{(n)} \cdot L_{seg}^{(n)}(\hat{y}^{s+(n)}, y^s) \\
 & + \lambda_{adv}^{(n)} \cdot L_{adv}^{DHA(n)}(\hat{z}^{s+(n)}, \hat{z}^{t+(n)}) \\
 & + \lambda_{distill}^{(n)} \cdot L_{distill}^{(n)}(\hat{y}^{s(n)}, y_{old}^{s(n)}) \Big],
\end{aligned}
\label{eq:ETMadapt1}
\end{equation}
\begin{equation}
\min \; \Big[L_D^{DHA(1)}(\hat{z}^{s+(1)}, \hat{z}^{t+(1)})\Big],
\label{eq:ETMadapt2}
\end{equation}
\begin{equation}
\min \; \Big[L_D^{DHA(2)}(\hat{z}^{s+(2)}, \hat{z}^{t+(2)})\Big].
\label{eq:ETMadapt3}
\end{equation}
The parameters of the ResNet 101, ASPP (1), and ASPP (2) modules are updated by Eq. (\ref{eq:ETMadapt1}), and the parameters of the Discriminators (1) and (2) are updated by Eq. (\ref{eq:ETMadapt2}) and (\ref{eq:ETMadapt3}), respectively.

\section{Implementation Details}
We used the SGD optimizer for the segmentation network and the TM. The learning rate was set to $2.5\times10^{-3}$ with the momentum of 0.97 and the weight decay value of $5\times10^{-4}$. Since we used the pre-trained ResNet 101 module, the learning rate of it was set 10 times smaller. To regularize the large shift of the previously learned parameters, the learning rate of the ASPP modules was set 10 times smaller, except when conducting the domain adaptation on the first target domain. On the other hand, we used Adam optimizer for the discriminator. The learning rate was set to $1\times10^{-4}$. We used the batch size of $1$ for the input size H and $3$ for the input size L for both source and target domain data. The temperature value for the distillation loss was set to 2. Additional hyperparameter values are specified in Table \ref{table:hyperparameters}. 

\begin{table}[h!]
\tiny
\centering
\renewcommand{\arraystretch}{1.3}
\caption{Hyperparameter values used in Eq. (\ref{eq:ETMadapt1})}
\label{table:hyperparameters}
\begin{tabular}{c|cccccc}
\hline
Hyperparameters & $\lambda_{seg}^{(1)}$ & $\lambda_{seg}^{(2)}$ & $\lambda_{adv}^{(1)}$ & $\lambda_{adv}^{(2)}$ & $\lambda_{distill}^{(1)}$ & $\lambda_{distill}^{(2)}$ \\
\hline
Values & 0.1 & 1 & 0.0002 & 0.001 & 0.02 & 0.2 \\
\hline
\end{tabular}
\end{table}

\section{Additional Experimental Results}
\paragraph{Qualitative Results.}
\begin{figure*}[!t]
\centering  
\subfigure[]{\includegraphics[width=0.979\linewidth]{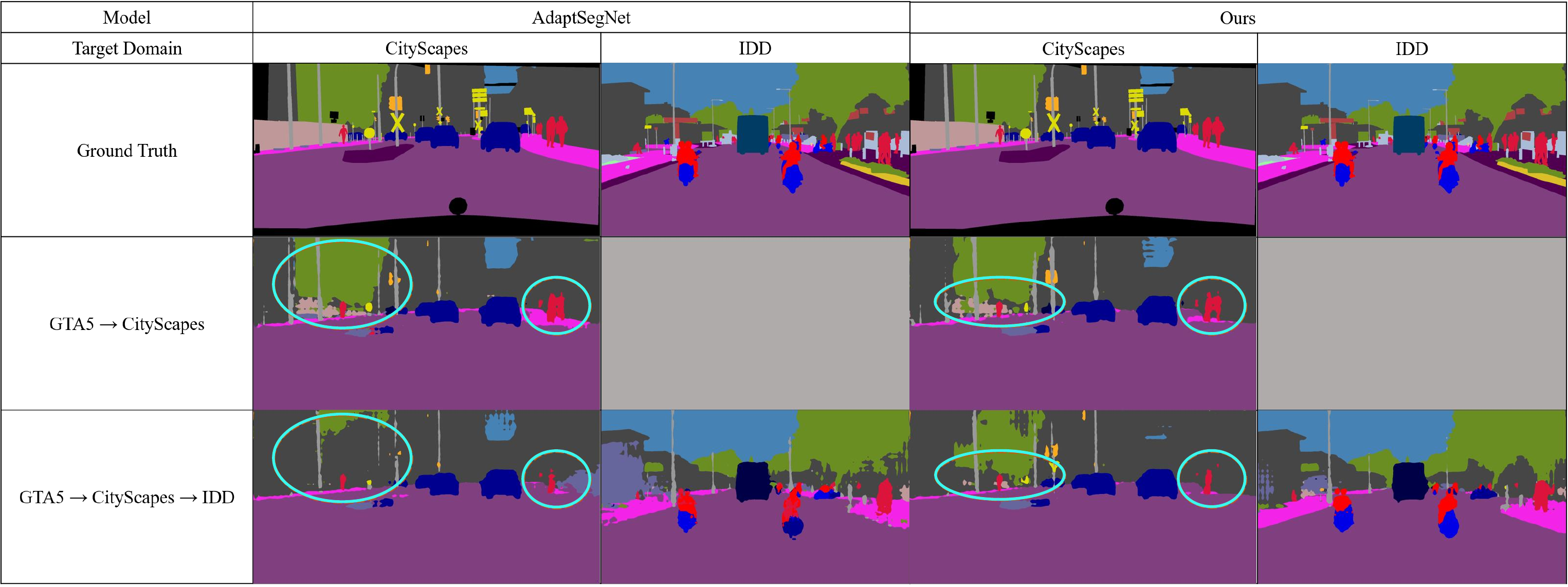}}
\subfigure[]{\includegraphics[width=0.9785\linewidth]{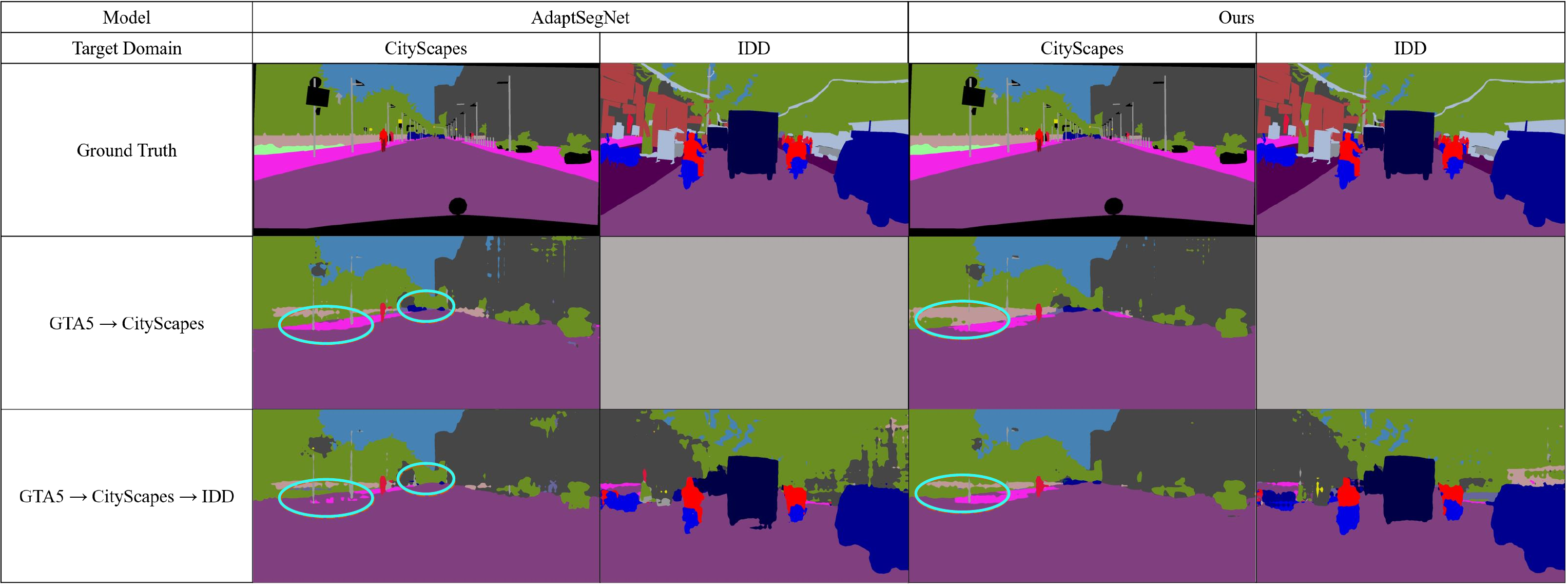}}
\caption{Qualitative results of Continual UDA on the CityScapes and IDD datasets from the GTA5 dataset. Comparing the AdaptSegNet and the model trained with our framework, less forgetting occurs when our framework is applied. Note that when UDA is done up to the CityScapese dataset, the semantic maps for the IDD dataset are meaningless (gray areas).}
\label{fig:qual}
\end{figure*}

To show qualitatively that the catastrophic forgetting problem is alleviated when using the proposed framework, semantic maps that the network predicted are provided in Fig. \ref{fig:qual}. Each semantic map is obtained through Continual UDA for the CityScapes and IDD datasets when the input size is H and the GTA5 dataset is used as the source domain.

In Fig. \ref{fig:qual}(a), the right parts are the examples of semantic maps predicted by the network with the ETM framework. The CityScapes column shows the semantic maps for an image extracted from the CityScapes dataset, and the IDD column shows the semantic maps for the image from the IDD dataset. In the first row (Ground Truth), we represent labels of each image, i.e., real semantic maps. In the second row (GTA5 $\rightarrow$ CityScapes), there are the semantic maps predicted by the network when it performs UDA up to the CityScapes dataset. Therefore, there are no semantic maps for the IDD dataset. In the last row (GTA5 $\rightarrow$ CityScapes $\rightarrow$ IDD), we represent the predicted semantic maps for the images extracted from each dataset after the network performing UDA up to the IDD dataset. Note that there is not much difference when comparing the second and third rows of the CityScapes columns. In other words, there is little forgetting on the CityScapes dataset even after learning about the IDD dataset continuously. Although the forgetting occurs for walls (expressed in beige) and people (expressed in red) inside the sky-blue circles, it is not serious.

\begin{table*}[!t]
\centering
\scriptsize
\caption{The Difference in the Performance of Semantic Segmentation by Objects with or without the TM}
\label{table:TMablation}
\begin{tabular}{@{}c|c||>{\centering\arraybackslash}p{0.97cm}>{\centering\arraybackslash}p{0.97cm}>{\centering\arraybackslash}p{0.97cm}>{\centering\arraybackslash}p{0.97cm}>{\centering\arraybackslash}p{0.97cm}>{\centering\arraybackslash}p{0.97cm}>{\centering\arraybackslash}p{0.97cm}>{\centering\arraybackslash}p{0.97cm}>{\centering\arraybackslash}p{0.97cm}||c@{}}
\toprule
\multicolumn{2}{c||}{\multirow{2}{*}{\textbf{Object}}}                    & \textbf{Road} & \textbf{Sidewalk} & \textbf{Building} & \textbf{Wall} & \textbf{Fence} & \textbf{Pole} & \textbf{\scalebox{.78}{Traffic Light}} & \textbf{\scalebox{.78}{Traffic Sign}} & \textbf{\scalebox{.78}{Vegetation}} & \multirow{2}{*}{\textbf{mIoU}}  \\
\multicolumn{2}{c||}{}                                                    & \textbf{Sky}  & \textbf{Person}   & \textbf{Rider}    & \textbf{Car}  & \textbf{Truck} & \textbf{Bus}  & \textbf{Train}         & \textbf{\scalebox{.78}{Motorcycle}}   & \textbf{Bicycle}    &                                 \\ \midrule
\multirow{4}{*}{\begin{tabular}[c]{@{}l@{}}\textbf{GTA5}\\ \textbf{$\rightarrow$ CityScapes}\end{tabular}}      & \multirow{2}{*}{\textbf{w/ TM}}  & 87.47         & 18.74             & 80.44             & 27.85         & 19.50          & 27.69         & 28.77                  & 16.45                 & 84.31               & \multirow{2}{*}{\textbf{41.96}} \\
                                      &                                  & 76.92         & 55.92             & 24.86             & 71.67         & 26.46          & 36.64         & 5.19                   & 24.38                 & 15.86               &                                 \\ \cmidrule(l){2-12} 
                                      & \multirow{2}{*}{\textbf{w/o TM}} & 86.33         & 16.38             & 76.40             & 24.42         & 16.53          & 21.92         & 4.26                   & 5.84                  & 80.31               & \multirow{2}{*}{\textbf{35.58}} \\
                                      &                                  & 75.06         & 37.71             & 17.68             & 69.37         & 26.18          & 36.64         & 5.19                   & 24.38                 & 15.86               &                                 \\ \midrule
\multirow{4}{*}{\begin{tabular}[c]{@{}l@{}}\textbf{GTA5}\\ \textbf{$\rightarrow$ CityScapes} \\ \textbf{$\rightarrow$ IDD}\end{tabular}} & \multirow{2}{*}{\textbf{w/ TM}}  & 94.95         & 43.18             & 55.31             & 31.12         & 21.57          & 24.70         & 9.02                   & 54.92                 & 83.32               & \multirow{2}{*}{\textbf{46.73}} \\
                                      &                                  & 92.94         & 40.47             & 46.65             & 74.01         & 59.76          & 40.978         & 0.12                   & 47.39                 & 20.68               &                                 \\ \cmidrule(l){2-12} 
                                      & \multirow{2}{*}{\textbf{w/o TM}} & 92.91         & 37.29             & 52.96             & 28.60         & 18.29          & 12.90         & 2.67                   & 49.80                 & 81.83               & \multirow{2}{*}{\textbf{43.42}} \\
                                      &                                  & 93.65         & 30.52             & 40.07             & 73.58         & 58.28          & 44.10         & 0.00                   & 41.69                 & 22.34               &                                 \\ \bottomrule
\end{tabular}
\end{table*}

The left parts are semantic maps predicted by the AdaptSegNet for the same situation and images. Unlike in the case of ours, the forgetting is more severe when comparing the second and third rows of the CityScapes column. Inside the sky-blue circle, one can see that there are forgettings on walls, vegetation (expressed in green), traffic lights (expressed in orange), sidewalks (expressed in light pink), and people. Furthermore, the model with the ETM framework predicts slightly better also for the IDD dataset. 

Fig. \ref{fig:qual}(b) shows the semantic maps for other examples. Overall, it shows the same tendency as in Fig. \ref{fig:qual}(a). When comparing the semantic maps for the CityScapes dataset, more severe forgetting occurs in the AdaptSegNet. Inside the sky-blue circle, there are forgettings on the walls, sidewalks, and vehicles (expressed in navy). Meanwhile, there are fewer forgetting phenomena in the ETM framework.

\paragraph{Further Analysis on the DHA Loss.}
\begin{table}[!t]
\centering
\tiny
\caption{The Continual UDA results according to the loss function for adversarial learning}
\label{table:DHAablation}

\begin{tabular}{@{}c||ccc||ccc@{}}
\toprule
\multirow{4}{*}{\begin{tabular}[c]{@{}c@{}}\textbf{Loss for}\\ \textbf{Adversarial Learning}\end{tabular}}
& \multicolumn{3}{c||}{\textbf{GTA5 $\rightarrow$ CityScapes $\rightarrow$ IDD}} 
& \multicolumn{3}{c}{\textbf{SYNTHIA $\rightarrow$ CityScapes $\rightarrow$ IDD}} \\ \cmidrule(l){2-7} 

& \textbf{CityScapes} & \textbf{IDD} 
& \begin{tabular}[c]{@{}c@{}}\textbf{Mean}\\ \textbf{mIoU}\end{tabular} 
& \textbf{CityScapes} & \textbf{IDD} 
& \begin{tabular}[c]{@{}c@{}}\textbf{Mean}\\ \textbf{mIoU}\end{tabular} \\ \midrule

\multicolumn{1}{c||}{\textbf{Loss\textsubscript{GAN}}}
& \underline{40.04} & \underline{44.28} & \underline{42.16} 
& \underline{43.56} & \underline{36.52} & \underline{40.04}  \\ \midrule

\multicolumn{1}{c||}{\textbf{Loss\textsubscript{GeoGAN}}}
& 36.50 & 40.35 & 38.43 
& 43.03 & 35.37 & 39.20  \\ \midrule

\multicolumn{1}{c||}{\textbf{DHA}}
& \textbf{40.61}  & \textbf{46.73}   & \textbf{43.67}  
& \textbf{43.86}  & \textbf{37.17}   & \textbf{40.52} \\ \bottomrule
\end{tabular}
\end{table}

To perform UDA with adversarial learning, the adversarial loss and discriminator loss are needed. Generally, the loss functions proposed in GAN \cite{goodfellow2014generative} are used. In our framework, the DHA loss inspired by Geometric GAN \cite{lim2017geometric} is utilized instead. Therefore, to validate the effectiveness of the proposed DHA loss, we conducted the experiments under the same conditions but with different loss functions for adversarial learning. In other words, only the adversarial loss and discriminator loss are changed while the other parts of the ETM framework remain the same, including the TM. In Table \ref{table:DHAablation}, the Continual UDA performances are reported when different loss functions are used: the loss functions from GAN and Geometric GAN, and the DHA loss. First of all, when the DHA loss is used, the model outperforms the other two models. Also noteworthy, the performance when with the GAN’s loss function is better than when the Geometric GAN’s loss is utilized. It shows a different tendency from the image generation field in which the Geometric GAN outperforms GAN. Therefore, it can be concluded that the loss function from the Geometric GAN is not suitable for UDA; however, the DHA loss, which is 
modified from the Geometric GAN’s loss, is suitable for UDA.

\paragraph{Further Analysis on the TM.}

Here, we present quantitative results to support which information that the TM contains (see Table \ref{table:TMablation}). In the upper row of the table (GTA5 $\rightarrow$ CityScapes), each object’s IoU and mIoU values for the CityScapes dataset are specified when UDA is performed to the CityScapes dataset with the GTA5 dataset as the source domain. In the lower row (GTA5 $\rightarrow$ CityScapes $\rightarrow$ IDD), the performance on the IDD dataset when continual UDA is done to the IDD dataset is reported. Overall, the IoU values are decreased for all objects without the TM compared to when the TM is used, and it leads to a reduction in the mIoU. Among the objects, the IoU values are decreased significantly for objects such as poles, traffic lights, traffic signs, and people. In other words, the TM contains much information about such objects. Through the experiment, it can be argued that the $1\times1$ Conv. module allows containing information about the instance objects and the Avg. Pool. module allows containing bits of information about all objects in general.

\end{document}